\documentclass{article}
\pdfoutput=1
\usepackage{microtype}
\usepackage{graphicx}
\usepackage{booktabs}
\usepackage{enumitem}
\usepackage{subcaption}
\usepackage{wrapfig}

\usepackage{hyperref}
\usepackage{multirow} 
\usepackage[dvipsnames,table]{xcolor}

\definecolor{darkgreen}{RGB}{0, 200, 0}

\usepackage[accepted]{icml2026}

\usepackage{amsmath}
\usepackage{amssymb}
\usepackage{pifont}
\usepackage{mathtools}
\usepackage{amsthm}
\usepackage{cuted}
\usepackage{enumitem}

\usepackage[capitalize,noabbrev]{cleveref}
\crefname{section}{Appendix}{Appendices}

\theoremstyle{plain}

\theoremstyle{definition}

\theoremstyle{remark}

\newcommand{\greencheck}{{\color{darkgreen}\checkmark}}
\newcommand{\redx}{{\color{red}\ding{55}}}

\definecolor{DustyRose}{RGB}{185, 90, 100}

\hypersetup{
    colorlinks,
    linkcolor={DustyRose}, 
    citecolor={DustyRose}, 
    urlcolor={DustyRose}
}
\usepackage[disable,textsize=tiny]{todonotes}
\def\pz{{\phantom{0}}}

\icmltitlerunning{Multimodal Crystal Flow: Any-to-Any Modality Generation for Unified Crystal Modeling
}

\begin{document}    

\twocolumn[
\icmltitle{Multimodal Crystal Flow: Any-to-Any Modality Generation \\ for Unified Crystal Modeling
}

\icmlsetsymbol{equal}{*}

\begin{icmlauthorlist}
\icmlauthor{Kiyoung Seong}{yyy,comp}
\icmlauthor{Sungsoo Ahn}{yyy}
\icmlauthor{Sehui Han}{comp}
\icmlauthor{Changyoung Park}{comp}
\end{icmlauthorlist}

\icmlaffiliation{yyy}{Graduate School of AI, KAIST, Seoul, South Korea}
\icmlaffiliation{comp}{Materials Intelligence Lab, LG AI Research, Seoul, South Korea}

\icmlcorrespondingauthor{Changyoung Park}{changyoung.park@lgresearch.ai}

\icmlkeywords{Machine Learning, ICML}

\vskip 0.3in
]

\printAffiliationsAndNotice{}

\begin{abstract}
Crystal modeling spans a family of conditional and unconditional generation tasks, including crystal structure prediction (CSP) and \emph{de novo} generation (DNG). While recent deep generative models have shown promising performance, they remain largely task-specific, lacking a unified framework that shares crystal representations across tasks. To address this limitation, we propose \emph{Multimodal Crystal Flow (MCFlow)}, a unified multimodal flow model that realizes multiple crystal generation tasks as distinct inference trajectories via independent time variables for atom types and crystal structures. To enable multimodal flow in a standard transformer model, we introduce a composition- and symmetry-aware atom ordering with hierarchical permutation augmentation, injecting compositional and crystallographic priors without explicit structural templates. Experiments on the MP-20 and MPTS-52 benchmarks show that a single MCFlow model is competitive with task-specific baselines across CSP, DNG, and structure-conditioned atom type generation.
\end{abstract}

\section{Introduction}

\begin{figure*}[t]
    \centering
    \begin{minipage}{0.48\textwidth}
        \centering
        \includegraphics[width=\linewidth]{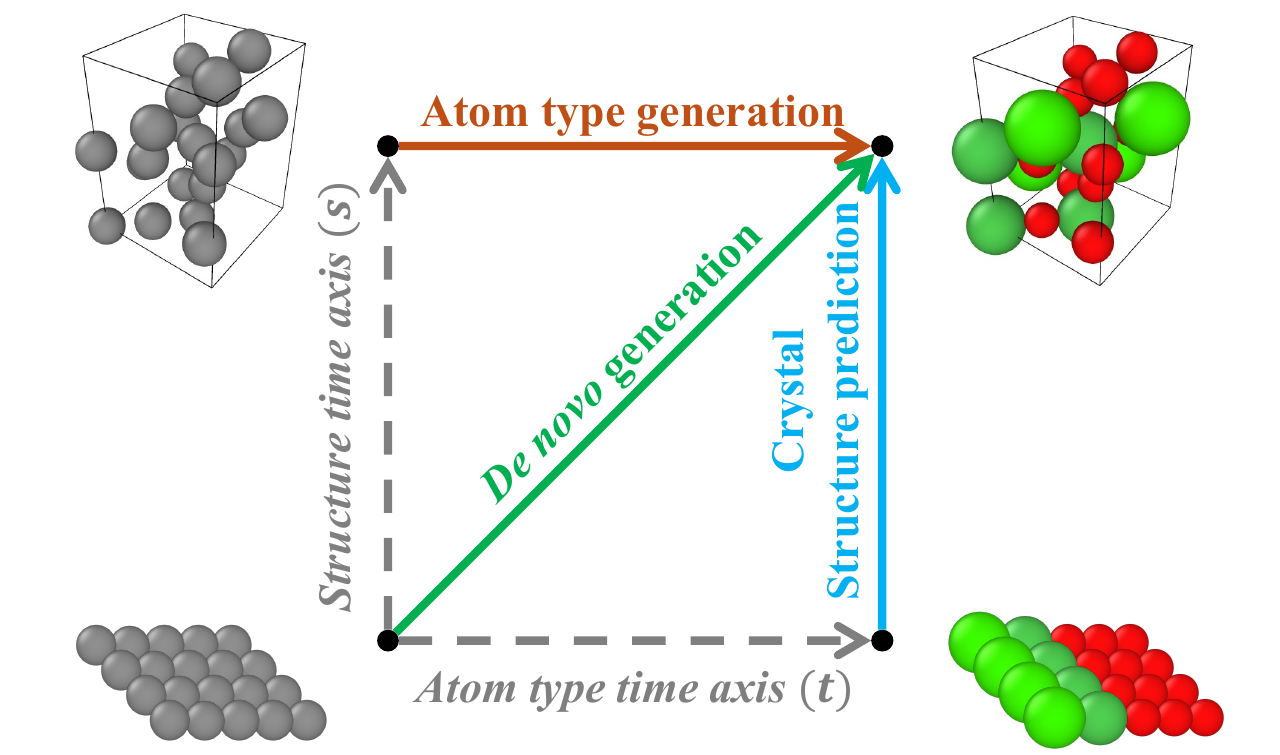}
        \label{fig:mcflow}
    \end{minipage}
    \hfill
    \begin{minipage}{0.48\textwidth}
        \centering
        \includegraphics[width=\linewidth]{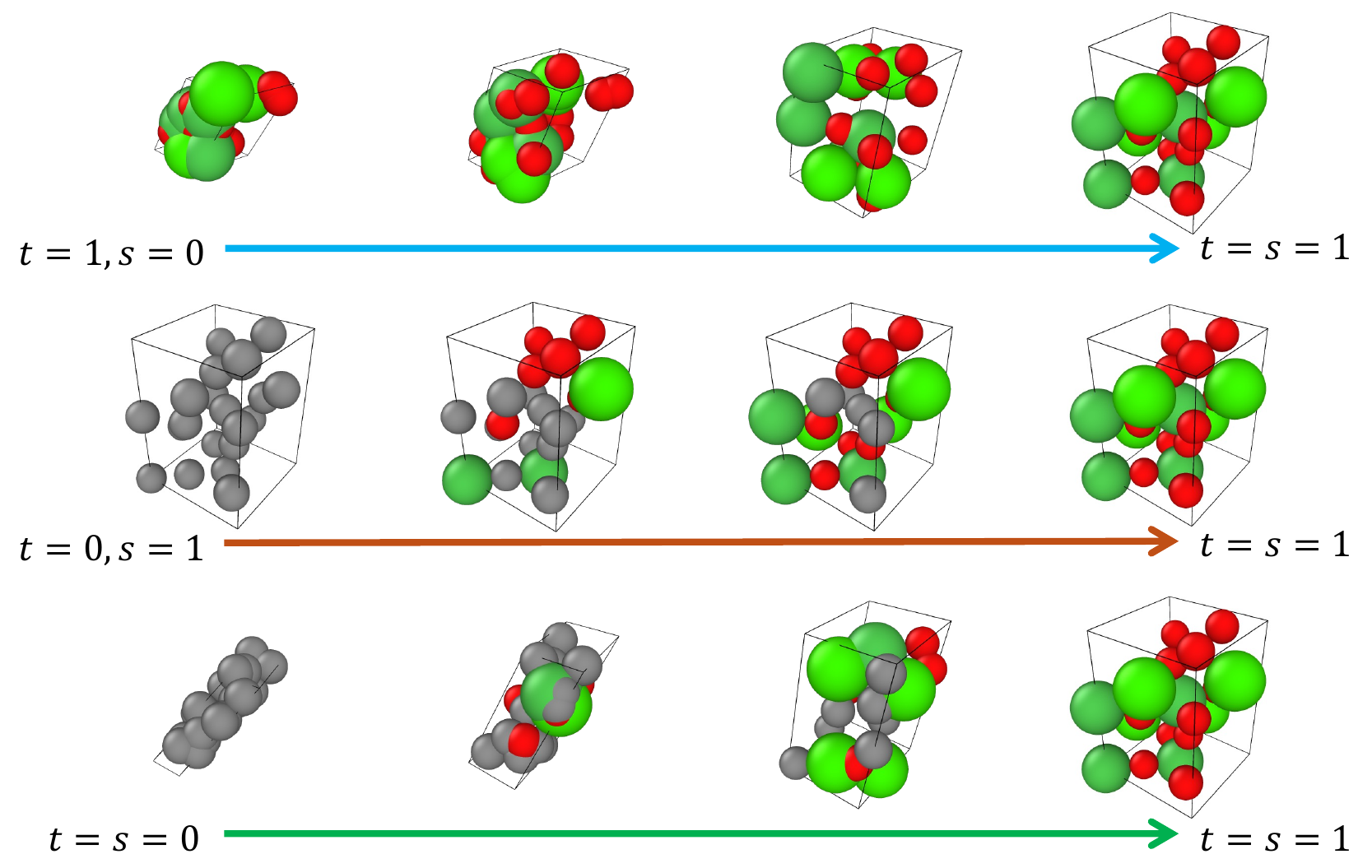}
        \label{fig:trajs}
    \end{minipage}

    \vspace{-15pt} 
    \caption{\textbf{Overview of multimodal crystal flow with any-to-any modality generation.} MCFlow trains a flow model with two independent time variables corresponding to atom types ($t$) and structures ($s$). By selecting task-specific inference trajectories in the $(t,s)$ space, a single model performs crystal structure prediction, atom type generation, and \emph{de novo} generation.}
    \label{fig:overview}
\end{figure*}

Crystal discovery based on atomic structures is vital for developing next-generation functional materials such as battery electrolytes and catalysts~\citep{butler2018machine}. However, traditional approaches are computationally prohibitive due to expensive density functional theory (DFT) calculations~\citep{glass2006uspex,wang2010crystal,pickard2011ab,wang2012calypso}. To address this issue, prior work has introduced generative models that learn distributions of stable crystals from data under various conditional or unconditional generation schemes~\citep{nouira2018crystalgan,noh2019inverse,xie2021crystal,jiao2023crystal,miller2024flowmm}. Recent methods have further incorporated crystallographic symmetry through structural templates such as Wyckoff positions~\citep{zhu2024wycryst,jiao2024space}.

At a high level, crystal modeling spans a family of generation problems across different modalities such as atom types and crystal structures. Depending on which modalities are observed, this formulation naturally covers crystal structure prediction (CSP), \emph{de novo} generation (DNG), and structure-conditioned atom type generation. This perspective aligns with a broader trend in generative modeling toward unified frameworks where a single model supports multiple conditional generation tasks across modalities~\citep{campbell2024generative,rojas2025diffuse}. However, existing crystal generative models adopt task-specific formulations, necessitating separate training for different generation settings and hindering a unified formulation that reuses the same underlying crystal representations across tasks.

In this work, we introduce \emph{Multimodal Crystal Flow (MCFlow)}, a unified generative model that enables any-to-any modality generation across atom types and crystal structures (\cref{fig:overview}). The key insight is that by decoupling flow time axes for different modalities, a single model can flexibly condition on any subset of modalities while generating the rest, without retraining or task-specific architectures. This design allows a single MCFlow model to support CSP, DNG, and atom type generation.

We instantiate MCFlow using a diffusion transformer (DiT) architecture~\citep{peebles2023scalable}, following recent successes in all-atom molecular and crystalline modeling~\citep{joshi2025all,frank2025sampling,yi2025crystaldit,kim2026catflow,kim2026atommof,kim2026flexible,morehead2026zatom}. However, the DiT lacks inductive biases for crystallographic symmetry and permutation invariance. To address this, we introduce a composition- and symmetry-aware atom ordering and hierarchical permutation augmentation. By lexicographically sorting atoms based on electronegativity and Wyckoff positions, we enable the DiT to internalize chemical and crystallographic structure from data without relying on explicit structural templates. The hierarchical permutation augmentation then enforces invariance across and within symmetry-equivalent orbits, providing a structured inductive bias over compositionally and crystallographically equivalent atoms.

Our main contributions are as follows:
\begin{itemize}
    \item We introduce \emph{MCFlow}, a unified multimodal flow framework for crystal generation that enables any-to-any modality generation across atom types and crystal structures within a single model.
    \item We propose a composition- and symmetry-aware atom ordering with hierarchical inter- and intra-orbit permutation augmentation, enabling DiT architectures to exploit compositional and crystallographic priors from data without explicit symmetry templates.
    \item We demonstrate that a single MCFlow model achieves competitive performance across crystal structure prediction, \emph{de novo} generation, and structure-conditioned atom type generation.
    \item Building on structure-conditioned atom type generation, we further enable atom type substitution by fixing atom types at known sites and sampling target sites from the same trained model.
\end{itemize}

\textbf{Conflict of Interest Disclosure.} Authors S. Han and C. Park are employed by LG AI Research, and K. Seong conducted this research during an internship at LG AI Research. MCFlow, the model proposed and evaluated in this paper, was developed at LG AI Research.

\section{Related Work}
\label{sec:related_work}

\textbf{Crystal generative models.}
Recent advances in diffusion- and flow-based generative models have significantly improved the fidelity and stability of generated crystal structures.
Diffusion models~\citep{xie2021crystal,jiao2023crystal,yang2023scalable,zeni2023mattergen} model crystal generation as iterative denoising of atomic coordinates in a primitive unit cell.
More recent work has explored flow-based models~\citep{miller2024flowmm,sriram2024flowllm,luo2025crystalflow}, stochastic interpolants~\citep{hoellmer2025omat}, periodic Bayesian flows~\citep{wu2025crysbfn}, and generative flow networks~\citep{crystalgfn2025}.
ADiT~\citep{joshi2025all} and Zatom-1~\citep{morehead2026zatom} use transformer architectures for unified all-atom modeling of molecules and crystals.
Large language models have also been adapted for crystal generation by fine-tuning on CIF-formatted data~\citep{gruver2024fine,mishra2025llamat} or incorporating text guidance and hybrid LLM-diffusion pipelines~\citep{ding2024matexpert,park2025chemeleon,das2025tgdmat,khastagir2025crysllmgen}.
However, these models generally lack explicit inductive biases for space-group symmetry and Wyckoff structure.

Alternative approaches enforce symmetry by generating structural templates via autoregressive Wyckoff position prediction~\citep{zhu2024wycryst,crystalformer2025,kazeev2025wyckoff,kelvinius2025wyckoffdiff,xu2025plaidpp} or restricting generation to the asymmetric unit (ASU)~\citep{jiao2024space,levy2025symmcd,chang2025space,puny2025space} conditioned on structural templates. While these methods satisfy crystallographic constraints, they are tailored to specific conditional generation settings and often rely on task-dependent structural template generation or retrieval modules. In contrast, MCFlow handles CSP, DNG, and atom type generation in a single end-to-end model by incorporating crystallographic symmetry as a soft inductive bias without structural template modules.

\textbf{Any-to-any multimodal generative models.}
Recent diffusion- and flow-based generative frameworks increasingly adopt unified multimodal formulations to jointly model heterogeneous modalities. By decoupling modality-wise time axes, these frameworks enable flexible conditional generation, including any-to-any settings. 
Multiflow~\citep{campbell2024generative} applies this paradigm to jointly model amino acid sequences and structures for protein co-design, while \citet{rojas2025diffuse} generalize multimodal diffusion models to arbitrary state spaces with applications to text--image domains. 
Despite their success, extending such frameworks to crystalline materials remains nontrivial due to periodic boundary conditions and crystallographic symmetry. 
MCFlow extends this multimodal generative paradigm to periodic crystal systems by decoupling atom types and crystal structures with modality-wise time axes, enabling a unified framework that supports CSP, DNG, and atom type generation without task-specific architectures or retraining.

\section{Method}
\label{sec:method}
In this section, we present a multimodal flow framework for unified crystal generation that supports flexible conditional generation across atom types and crystal structures within a single model. The core idea is to decouple the generative processes of different modalities through modality-specific time parameterization, allowing diverse crystal generation tasks to be realized within a shared model. Building on this formulation, we introduce a composition- and symmetry-aware atom ordering together with permutation augmentation as a soft inductive bias for transformer-based architectures. Background on crystal representations and flow models is provided in \cref{app:background}.

\subsection{Multimodal Flow Model for Crystals}
\label{subsec:multimodal_flow}

Our goal is to generate arbitrary subsets of modalities while conditioning on the others. Standard diffusion or flow models rely on a single shared time variable, which couples all modalities to the same noise level and restricts conditional generation. We address this limitation by decoupling flow time variables to atom types and crystal structures and learning a joint generative process over a two-dimensional time space. Intuitively, this decoupling allows the model to hold one modality fixed while sampling the other, providing controllable conditional generation. Different crystal generation tasks then correspond to different inference trajectories in the $(t,s)$ plane (\cref{fig:overview}): CSP fixes $t=1$ and integrates $s$ from $0$ to $1$, atom type generation fixes $s=1$ and integrates $t$, and DNG integrates $(t,s)$ jointly along the diagonal.

\textbf{Multimodal flow model.} We adopt a multimodal flow model with decoupled flow time variables $t,s\in[0,1]$ corresponding to atom types $A\in\mathbb{Z}_{+}^{N}$ and crystal structures $X=(F,L^{l},L^{a})$ in the primitive unit cell, which consists of fractional coordinates $F\in[0,1]^{3\times N}$, Niggli-reduced lattice lengths $L^l=(a, b, c)\in\mathbb{R}^{1\times3}$, and angles $L^a=(\alpha, \beta, \gamma)\in \left[60^\circ,120^\circ\right]^{1\times3}$~\citep{grosse2004numerically}. After training the flow model, we select an inference trajectory in a $(t,s)$ time plane to realize different crystal discovery tasks with observed modalities fixed at $1$.

To describe the multimodal flow model, we first design the joint probability path $p_{t,s}(A_t,X_s)$ defined by marginalizing modality-wise conditional paths:
\begin{equation}
\begin{aligned}
p_{t,s}(A_t,X_s)
&= \int p_t(A_t \mid A_0, A_1) \\
&\qquad p_s(X_s \mid X_0, X_1)\, d\pi,
\end{aligned}
\end{equation}
where $\pi(A_0,X_0,A_1,X_1)=p_{0}(A_0,X_0)p_{1}(A_1,X_1)$ is the independent coupling between joint base $p_0$ and data $p_1$ distributions over atom types and structures. By construction, this path recovers the joint base $p_0$ at $(t,s)=(0,0)$ and the joint data $p_1$ at $(t,s)=(1,1)$, interpolating between them through independent time evolution along each modality. 

This joint path evolves through a continuous-time Markov chain (CTMC) over atom types along the $t$-axis and an ordinary differential equation (ODE) over crystal structures along the $s$-axis, expressed as:
\begin{equation}
\begin{aligned}
p_{t+h\mid t,s}(\cdot\mid A_t,X_s)
&= \delta_{A_t} + h\,u^\text{A}(A_t,X_s,t,s) \\
&\quad + o(h),
\end{aligned}
\end{equation}
\begin{equation}
    \frac{dX_s}{ds} = u^\text{X}(A_t,X_s,t,s),
\end{equation}
where $\delta_{A_t}$ denotes the probability vector supported on the current state $A_t$, $u^\text{A}(A_t,X_s,t,s)$ denotes the rate vector of the CTMC over atom types in a discrete state space, $u^\text{X}(A_t,X_s,t,s)$ denotes the velocity field of the ODE over crystal structures in a continuous state space, and $o(h)$ denotes residual terms that vanish as $h \to 0$. The $t$-axis governs atom type updates conditioned on crystal structures, whereas the $s$-axis handles crystal structure updates conditioned on atom types.

\begin{figure*}[t]
    \centering
    \includegraphics[width=\textwidth]{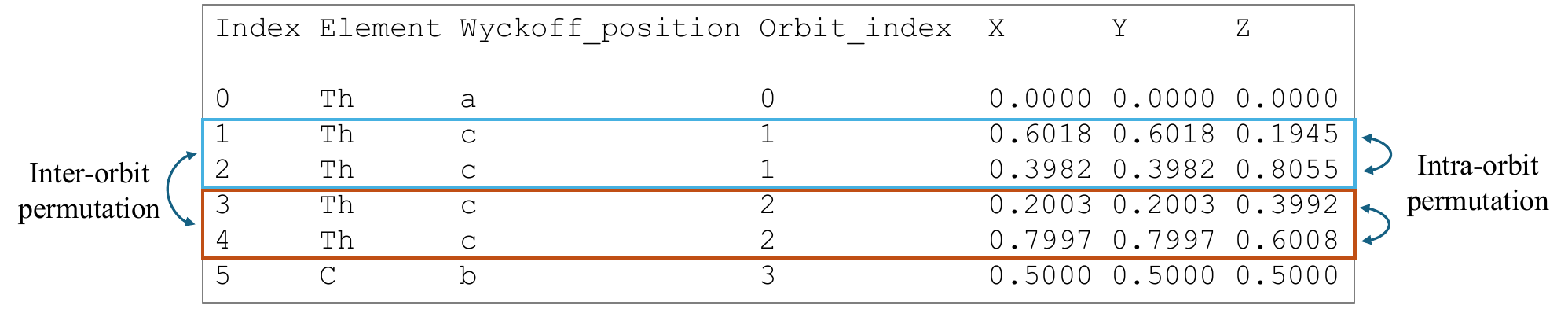}
    \vspace{-15pt}
    \caption{
    \textbf{Composition- and symmetry-aware atom ordering with hierarchical permutation augmentation.}
    Atoms in the primitive unit cell are lexicographically ordered by Pauling electronegativity and Wyckoff position (denoted by letter $a,b,c,\ldots$) to expose compositional and crystallographic structure.
    The ordering and augmentation are illustrated on a $\mathrm{Th}_5\mathrm{C}$ crystal in the $\mathrm{R}\bar{3}\mathrm{m}$ space group.
    Inter-orbit permutation reorders orbits sharing the same elements and Wyckoff position, while intra-orbit permutation permutes atoms within each orbit.
    This ordering and hierarchical permutation augmentation provides compositional and crystallographic symmetry information to sequence-based Transformers without explicitly conditioning space group or Wyckoff positions.
    }
    \label{fig:ordering}
\end{figure*}

\textbf{Multimodal flow matching.}
We adopt a flow matching objective that minimizes a weighted sum of divergences between the model prediction and the modality-specific target velocity field or rate vector:
\begin{equation}
\mathcal{L}(\theta)=\mathbb{E}_{t,s,A_t,X_s} \left[ \sum_{m \in \{\text{A}, \text{X}\}} w_m D_m(u^m_\theta, u^m) \right], 
\end{equation}
where $u^m_\theta \coloneqq u^m_\theta(A_t, X_s, t, s)$ and $u^m \coloneqq u^m(A_t, X_s, t, s)$, $D_\text{A}$ is generalized KL divergence for atom types and $D_\text{X}$ is squared Euclidean distance for crystal structures.

For tractable supervision, we match to modality-wise conditional targets $u^\text{A}(A_t \,|\, A_0,A_1)$ and $u^\text{X}(X_s \,|\, X_0,X_1)$ rather than directly to the marginal targets $u^\text{A}(A_t,X_s,t,s)$ and $u^\text{X}(A_t,X_s,t,s)$, since the marginal targets are the posterior expectation of the modality-wise conditional targets under endpoint coupling
\begingroup
\setlength{\abovedisplayskip}{12pt}
\setlength{\belowdisplayskip}{12pt}
\begin{align}
u^\text{A} &= \mathbb{E}_{A_0,A_1} \left[u^\text{A}(A_t \,|\, A_0,A_1)\,\vert\,A_t,X_s\right], \\
u^\text{X} &= \mathbb{E}_{X_0,X_1} \left[u^\text{X}(X_s \,|\, X_0,X_1)\,\vert\,A_t,X_s\right].
\end{align}
\endgroup
Under these divergences, the conditional flow matching objective induces the same gradient with respect to the model parameters as matching the marginal target implied by this posterior expectation~\citep{lipman2024flowmatchingguidecode}. We construct conditional target velocity fields or rate vector as follows: straight line flows between prior and target samples for lattice parameter, geodesic flows between prior and target samples on the torus manifold $\mathbb{T}^{3\times N}$ for fractional coordinates~\citep{miller2024flowmm} to enforce periodicity, and mixture paths between two point masses of prior and target samples for atom types; full expressions are provided in \cref{app:mcflow}.

\textbf{Parameterization and base distributions.}
We adopt clean data prediction across all modalities instead of predicting noise or velocity targets, motivated by the manifold structure of data distribution~\citep{li2025back}. For fractional coordinates, we predict fractional coordinates $F_\theta$ and project them onto the torus $\mathbb{T}^{3\times N}$ via $\text{proj}(F)=F \pmod 1$.
We use a uniform base distribution on $\mathbb{T}^{3\times N}$ for fractional coordinates.
For lattice length and angles, we predict directly $L^{l}_{\theta}$ and $L^{a}_{\theta}$, using LogNormal priors for lengths to ensure positivity~\citep{miller2024flowmm} and uniform priors for angles.
For atom types, we predict atom type logits $A_{\theta}$ with a mask-token point mass $\delta_M$ as base distribution.

\textbf{Diffusion transformer.}
For all-atom modeling of crystals, we adopt a Diffusion Transformer (DiT) backbone~\citep{peebles2023scalable} with modality-specific time conditioning and prediction heads tailored to the proposed multimodal flow. This choice follows recent successes of transformer architectures in all-atom molecular and crystalline modeling~\citep{joshi2025all,frank2025sampling,morehead2026zatom}, providing a scalable backbone that uniformly handles heterogeneous modalities through attention. We embed all modalities and add a fixed sinusoidal positional embedding $\text{PosEmb}$~\citep{vaswani2017attention} indexed by the atom position in the proposed atom ordering (Section~\ref{sec:symmetry_aug}) to obtain a token embedding:
\begin{equation}
\begin{aligned}
z = {} &\text{Emb}(A) + \text{Lin}(F) + \text{Lin}(L^l) + \text{Lin}(L^a) \\
       &+ \text{PosEmb},
\end{aligned}
\end{equation}
where $\text{Lin}(\cdot)$ denotes modality-specific linear layer and lattice embeddings are broadcasted to all atom tokens.
The joint conditioning vector for learnable layer normalization is a sum of modality-wise time embeddings
\begin{equation}
c=\text{TimeEmb}(t)+\text{TimeEmb}(s),
\end{equation}
where $\text{TimeEmb}(\cdot)$ denotes modality-specific time embedding layer.
We feed the multimodal embedding vector $z$ and the conditioning vector $c$ into the DiT backbone,
$h = \mathrm{DiT}(z, c)$.
From the token-wise representations, we linearly predict atom type logits and fractional coordinates,
\begin{equation}
A_{\theta} = \mathrm{Lin}(h), \quad F_{\theta} = \mathrm{Lin}(h).
\end{equation}
Global lattice parameters are predicted via mean pooling.
\begin{equation}
L^l_\theta = \mathrm{Lin}(\bar{h}), \quad
L^a_\theta = \mathrm{Lin}(\bar{h}), \quad
\bar{h} = \frac{1}{N}\sum_{i=1}^N h_i.
\end{equation}
The final velocity field and rate vector are constructed from these outputs as provided in \cref{app:mcflow}. Training and inference algorithms are listed in \cref{app:algorithm}. Since crystals have variable numbers of atoms, we pad sequences within a batch and use attention masks; the pooling for lattice heads is computed over non-padded atom tokens only.

\textbf{Inference-time guidance.}
To improve sample fidelity in our multimodal flow framework, we adopt noisy guidance~\citep[NG;][]{rojas2025diffuse}, originally introduced for diffusion score models, and apply it to the velocity field of multimodal flow as an inference-time technique for conditional generation. This follows from the probability flow ODE~\citep{song2020score}, where the score function and the velocity field are deterministically related. For crystal structure prediction, the guided velocity field is given by
\begin{equation}
\begin{aligned}
u^\text{X}_{\text{ng}}(A_1, X_s, 1, s)
&= (1-\omega)\,u^\text{X}(A_\sigma, X_s, \sigma, s) \\
&\quad + \omega\,u^\text{X}(A_1, X_s, 1, s).
\end{aligned}
\end{equation}
where $\sigma$ controls the corruption level of the conditioning modality and $\omega$ denotes the guidance scale, biasing inference trajectories toward the regions favored by the conditional velocity field. The noisy guidance procedure is summarized in \cref{alg:ng}, with additional ablation and sensitivity analyses provided in \cref{app:guidance}.

\subsection{Composition- and Symmetry-Aware Atom Ordering and Hierarchical Augmentation}
\label{sec:symmetry_aug}

This subsection introduces an atom ordering and augmentation strategy for the DiT architecture, designed to support multimodal flow models by incorporating compositional and crystallographic symmetry information.

Prior work has proposed ordering atoms in crystal structures according to their Wyckoff positions for an autoregressive transformer model, leveraging crystallographic symmetry~\citep{crystalformer2025}. While effective, directly applying Wyckoff-based ordering to a DiT-based multimodal flow model suffers from two limitations. First, in crystal structure prediction (CSP), where only the composition is given, the absence of structural symmetry information at inference time leads to ambiguity in defining the initial atom ordering for the DiT. Second, this ordering does not explicitly encode chemical information, which can limit the model's ability to learn compositionally valid crystals.

\textbf{Composition- and symmetry-aware atom ordering.}
To address these limitations, we propose a lexicographical atom ordering that jointly incorporates compositional
and crystallographic information. As illustrated in \cref{fig:ordering}, we first partition atoms into groups $O = \{ g \cdot f \mid g \in G \}$ under the action of the space group $G$, referred to as \emph{orbits}. Each orbit represents compositionally and symmetry-equivalent atomic sites sharing the same atom type and Wyckoff position.

We then impose a lexicographical ordering over these orbits, sorting them first by the Pauling electronegativity
of their atom type and then by Wyckoff position, as defined in the International Tables for Crystallography~\citep{prince2004international}. Electronegativity provides a
chemical signal that correlates with bonding preferences and charge-transfer
tendencies, while Wyckoff positions capture crystallographic symmetry with earlier letters indicating higher site symmetry. This ordering injects a soft inductive bias by arranging atoms in a symmetry-aligned and contiguous sequence, enabling the DiT to learn symmetry without explicit conditioning on space group or Wyckoff positions.

\textbf{Augmentation via inter- and intra-orbit permutation.}
While the proposed ordering captures compositional and symmetry information, permutation ambiguity remains
among tied orbits and atoms within the same orbit. We address this by introducing a hierarchical permutation
augmentation scheme that preserves compositional and crystallographic equivalence. For each group of orbits
$\mathcal{W}^i = (O^i_1, \dots, O^i_m)$ sharing the same atom type and Wyckoff position in a primitive unit cell, we apply inter-orbit
permutation across orbits followed by intra-orbit permutation within each orbit. This reduces the permutation space from $N!$ to $\prod_i |\mathcal{W}^i|! \prod_j |O^i_j|!$, with a quantitative analysis provided in
\cref{app:ordering}. We additionally apply a global modulo translation to account for translational invariance.

\begin{table}[t]
\centering
\caption{
\textbf{Crystal structure prediction on MP-20 and MPTS-52.}
MCFlow achieves the best results on MPTS-52 across all metrics and is competitive on MP-20, while achieving consistently lower RMSE.
We report match rate (MR) and RMSE for single-sample (1) and multi-sample (20) generation.
S/B/L denote small/base/large model sizes.
Noisy guidance is applied only for single-sample generation.
\textbf{Bold} indicates the best value within each column across all models and sampling budgets; \underline{underline} indicates the best value within the single-sample group.
}
\label{tab:csp}
\resizebox{\columnwidth}{!}{
\begin{tabular}{lccccc}
\toprule
\multirow{2}{*}{Model} & \multicolumn{2}{c}{MP-20} & \multicolumn{2}{c}{MPTS-52} \\
\cmidrule(lr){2-3} \cmidrule(lr){4-5}
& MR (\%) $\uparrow$ & RMSE $\downarrow$ & MR (\%) $\uparrow$ & RMSE $\downarrow$ \\
\midrule
CDVAE (1)       & 33.90 & 0.1045 & 5.34  & 0.2106 \\
DiffCSP (1)     & 51.49 & 0.0631 & 12.19 & 0.1786 \\
FlowMM (1)      & 61.39 & 0.0566 & 17.54 & 0.1726 \\
CrystalFlow (1) & 62.02 & 0.0710 & 21.00 & 0.1613 \\
OMatG (1)       & 63.75 & 0.0720 & 25.15 & 0.1931 \\
DiffCSP (20)    & 77.93 & 0.0492 & 34.02 & 0.1749 \\
CrystalFlow (20)& \textbf{78.34} & 0.0577 & 37.81 & 0.1584 \\
\midrule
MCFlow-S (1)  & 56.10 & 0.1269 & 22.88 & 0.1845 \\
MCFlow-B (1)  & 63.14 & 0.0611 & 26.46 & 0.1572 \\
MCFlow-L (1)  & \underline{64.08} & \underline{0.0561} & \underline{27.16} & \underline{0.1401} \\
MCFlow-S (20) & 77.84 & 0.0939 & 38.91 & 0.1717 \\
MCFlow-B (20) & 77.12 & 0.0439 & 41.40 & 0.1465 \\
MCFlow-L (20) & 76.08 & \textbf{0.0383} & \textbf{41.45} & \textbf{0.1296} \\
\bottomrule
\end{tabular}
}
\end{table}

\begin{table*}[t]
  \centering
  \caption{
  \textbf{De novo generation evaluated on LeMat-GenBench~\citep{lemat2025genbench} with MP-20 training.}
  MCFlow achieves competitive or superior performance across most metrics both with and without MLIP pre-relaxation.
  Reported values are averaged over an ensemble of three MLIPs (MACE-MP, UMA, Orb-v3) with LeMat-Bulk as the reference set, with all percentages computed over the 2{,}500 submitted structures following the LeMat-GenBench convention.
  Standard deviations are computed across the 2{,}500 generated structures, averaged over the three MLIPs.
  SUN and MSUN values are reported as percentage (count).
  Results are organized into two groups: models with and without MLIP pre-relaxation; MCFlow uses Orb-v3 for pre-relaxation.
  \textbf{Bold} and \underline{underline} indicate the best and second-best results within each group.
  }
  \label{tab:lemat}
  \resizebox{\textwidth}{!}{%
  \begin{tabular}{@{}lrrrrrrrrrrr@{}}
    \toprule
    \multirow{2}{*}{Model} & \multirow{2}{*}{\# Params} & \multirow{2}{*}{Valid (\%) $\uparrow$} & \multirow{2}{*}{Unique (\%) $\uparrow$} & \multirow{2}{*}{Novel (\%) $\uparrow$} & \multicolumn{3}{c}{Energy-based} & \multicolumn{2}{c}{Stability} & \multicolumn{2}{c}{Metastability} \\
    \cmidrule(lr){6-8} \cmidrule(lr){9-10} \cmidrule(lr){11-12}
    & & & & & $E_f$ (eV) $\downarrow$ & $E_{\text{hull}}$ (eV) $\downarrow$ & RMSD (\AA) $\downarrow$ & Stable (\%) $\uparrow$ & SUN (\%) $\uparrow$ & Meta (\%) $\uparrow$ & MSUN (\%) $\uparrow$ \\
    \midrule
    \textit{With pre-relaxation} \\
    MatterGen       & 47M  & 95.7 & 95.1 & \textbf{70.5} & $-0.70 \pm 0.79$ & $0.18 \pm 0.18$ & $0.39 \pm 0.50$ & 2.0  & 0.2 (6)  & 33.4 & \underline{15.0} \\
    PLaID++         & 7B   & 96.0 & 77.8 & 24.2          & $-0.50 \pm 0.44$ & $0.09 \pm 0.16$ & $0.13 \pm 0.29$ & 12.4 & \textbf{1.0 (26)} & 60.7 & 7.6 \\
    WyFormer        & 8B   & 93.4 & 93.0 & \underline{66.4} & $-0.43 \pm 0.95$ & $0.50 \pm 0.51$ & $0.81 \pm 0.98$ & 0.5  & 0.1 (3)  & 15.7 & 1.9 \\
    WyFormer-DFT    & 8B   & 95.2 & 95.0 & \underline{66.4} & $-0.67 \pm 0.91$ & $0.27 \pm 0.36$ & $0.42 \pm 0.60$ & 3.7  & 0.4 (9)  & 24.8 & 7.8 \\
    \midrule
    \rowcolor{DustyRose!10} MCFlow-S        & 34M  & 97.2 & \textbf{96.3} & 52.2 & $-0.85 \pm 0.84$ & $0.10 \pm 0.12$ & $0.16 \pm 0.27$ & 11.7 & \underline{0.7 (17)} & 49.5 & \textbf{18.9} \\
    \rowcolor{DustyRose!10} MCFlow-B        & 134M & \underline{97.7} & \underline{95.5} & 25.4 & $\underline{-0.91} \pm \underline{0.85}$ & $\underline{0.05} \pm \underline{0.10}$ & $\underline{0.08} \pm \underline{0.18}$ & \underline{17.6} & \underline{0.7 (17)} & \underline{64.3} & 11.9 \\
    \rowcolor{DustyRose!10} MCFlow-L        & 464M & \textbf{98.6} & 95.2 & 18.6 & $\mathbf{-0.93} \pm \mathbf{0.87}$ & $\mathbf{0.04} \pm \mathbf{0.08}$ & $\mathbf{0.06} \pm \mathbf{0.15}$ & \textbf{18.8} & 0.5 (13) & \textbf{68.3} & 9.3 \\
    \midrule
    \textit{Without pre-relaxation} \\
    ADiT            & 180M & 90.6 & 87.8 & 26.0          & $-0.73 \pm 0.93$ & $0.33 \pm 0.45$ & $0.38 \pm 0.40$ & 0.4  & 0.0 (0)  & 36.5 & 1.0 \\
    Crystal-GFN     & 1M   & 51.7 & 51.7 & 51.7          & $\mathbf{-1.30} \pm \mathbf{2.63}$ & $2.09 \pm 2.38$ & $1.87 \pm 0.97$ & 0.0  & 0.0 (0)  & 0.0  & 0.0 \\
    CrystalFormer   & 5M   & 69.9 & 69.4 & 31.8          & $-0.17 \pm 1.48$ & $0.70 \pm 1.33$ & $0.66 \pm 1.05$ & 1.4  & 0.0 (0)  & 28.8 & 3.1 \\
    DiffCSP         & 12M  & \underline{95.7} & \underline{94.8} & \textbf{66.2} & $-0.64 \pm 0.96$ & $0.28 \pm 0.56$ & $0.59 \pm 0.63$ & 2.3  & 0.1 (3)  & 29.8 & \textbf{8.5} \\
    DiffCSP++       & 12M  & 95.3 & \textbf{95.1} & \underline{62.0} & $-0.52 \pm 0.87$ & $0.41 \pm 0.44$ & $0.69 \pm 0.82$ & 1.0  & \textbf{0.2 (4)} & 26.4 & \underline{5.0} \\
    LLaMat2-CIF     & 7B   & 84.4 & 81.4 & 30.0          & $-0.47 \pm 1.01$ & $0.44 \pm 0.71$ & $0.54 \pm 0.71$ & 0.7  & 0.1 (3)  & 34.7 & 2.1 \\
    LLaMat3-CIF     & 8B   & 15.4 & 15.2 & 10.5          & $0.74 \pm 2.69$  & $1.71 \pm 2.48$ & $1.01 \pm 1.10$ & 0.1  & 0.0 (0)  & 2.1  & 0.2 \\
    SymmCD          & 12M  & 73.4 & 73.0 & 47.0          & $-0.02 \pm 1.22$ & $0.88 \pm 1.07$ & $0.87 \pm 1.09$ & 1.4  & 0.1 (2)  & 18.6 & 2.4 \\
    \midrule
    \rowcolor{DustyRose!10} MCFlow-S        & 34M  & 86.5 & 85.8 & 48.2 & $-0.42 \pm 0.95$ & $0.49 \pm 0.59$ & $0.64 \pm 0.61$ & 1.2  & 0.1 (2)  & 30.7 & 2.5 \\
    \rowcolor{DustyRose!10} MCFlow-B        & 134M & 93.3 & 91.2 & 25.1          & $-0.76 \pm 0.88$ & $\underline{0.18} \pm \underline{0.35}$ & $\underline{0.29} \pm \underline{0.44}$ & \underline{4.6} & \textbf{0.2 (4)} & \underline{60.4} & 3.2 \\
    \rowcolor{DustyRose!10} MCFlow-L        & 464M & \textbf{96.0} & 92.9 & 19.4 & $\underline{-0.84} \pm \underline{0.88}$ & $\mathbf{0.13} \pm \mathbf{0.27}$ & $\mathbf{0.23} \pm \mathbf{0.37}$ & \textbf{4.9} & 0.1 (2)  & \textbf{69.7} & 3.7 \\
    \bottomrule
  \end{tabular}%
  }
\end{table*}

\section{Experiments}
\label{sec:experiments}
We aim to answer two research questions:
\begin{itemize}[leftmargin=*]
    \item Can a single MCFlow model achieve competitive performance across crystal structure prediction (CSP), \emph{de novo} generation (DNG), and structure-conditioned atom type generation?
    \item Do the proposed composition- and symmetry-aware atom ordering and symmetry-preserving hierarchical permutation augmentations enable a transformer-based model to capture crystallographic constraints without relying on explicit symmetry templates? 
\end{itemize} 
To this end, we evaluate MCFlow on all three tasks using the same trained model and conduct ablation studies to assess the contribution of each component. Detailed experimental configurations and metrics are in \cref{app:config,app:metrics}.

\textbf{Dataset.} Our main evaluation uses two inorganic crystal benchmarks that support all three tasks (CSP, DNG, ATG): \emph{MP-20}~\citep{xie2021crystal} and \emph{MPTS-52}~\citep{baird2024mpts}. MP-20 comprises all materials from the Materials Project database~\citep{jain2013commentary} as of July 2021 that contain no more than 20 atoms per unit cell and reside within 0.08 eV/atom of the convex hull energy. We also test on MPTS-52, a dataset consisting of structures with up to 52 atoms per unit cell. For both datasets, we follow a consistent 60/20/20 ratio for training, validation, and test sets. MPTS-52 dataset is partitioned into time slices based on their initial publication year to reflect a chronological discovery.

\textbf{Setup.}
MCFlow is based on a Diffusion Transformer (DiT) backbone~\citep{peebles2023scalable} for all-atom crystal modeling. We evaluate three model scales: small (34M parameters, hidden dimension 384, 6 attention heads, 12 layers), base (134M parameters, 768 dimension, 12 heads, 12 layers), and large (464M parameters, 1024 dimension, 16 heads, 24 layers), denoted as MCFlow-S, -B, and -L, respectively. Throughout this paper, MCFlow refers to the base model unless its scale is explicitly stated. Detailed model, training, and inference configurations are provided in \cref{app:config}. Computational cost analysis is provided in \cref{tab:compute}.

\subsection{Crystal Structure Prediction}
CSP predicts stable crystal structures given a chemical composition: the number of atoms and their types.

\textbf{Metrics.} Following the metric calculation of \citet{jiao2023crystal,miller2024flowmm}, we evaluate the accuracy of the generated samples against the ground truth structures of the test set using the \texttt{StructureMatcher}~\citep{ong2013python}. We report the Match Rate (MR), defined as the fraction of valid candidates that match their ground truth counterparts. For matched pairs, we also calculate the Root Mean Square Error (RMSE) to assess the precision of the predicted atomic locations within the unit cell. We additionally report METRe~\citep{martirossyan2025metre} on a polymorph-aware MP-20 split, which accounts for compositions admitting multiple stable structures (\cref{tab:metre}).

\textbf{Baselines.} 
We compare MCFlow against baselines: CDVAE~\citep{xie2021crystal}, DiffCSP~\citep{jiao2023crystal}, FlowMM~\citep{miller2024flowmm}, CrystalFlow~\citep{luo2025crystalflow}, and OMatG~\citep{hoellmer2025omat}, for which we report the best-performing LinearODE variant. We exclude FlowLLM~\citep{sriram2024flowllm}, SymmCD~\citep{levy2025symmcd}, and ADiT~\citep{joshi2025all}, as they do not support CSP. We also exclude DiffCSP++~\citep{jiao2024space}, SGEquiDiff~\citep{chang2025space}, and SGFM~\citep{puny2025space}, which rely on structural templates extracted from the target structure or retrieved via CSPML~\citep{kusaba2022crystal}. For a given composition, multiple templates often exist, while one template leads to a nearly unique stable structure~\citep{kazeev2025wyckoff}, making template-conditioned generation a different problem setting from full CSP. We further compare our method under target-template filtering against these target template-conditioned methods (\cref{tab:csp_template}).

\begin{figure}[t]
    \centering
    \includegraphics[width=\columnwidth]{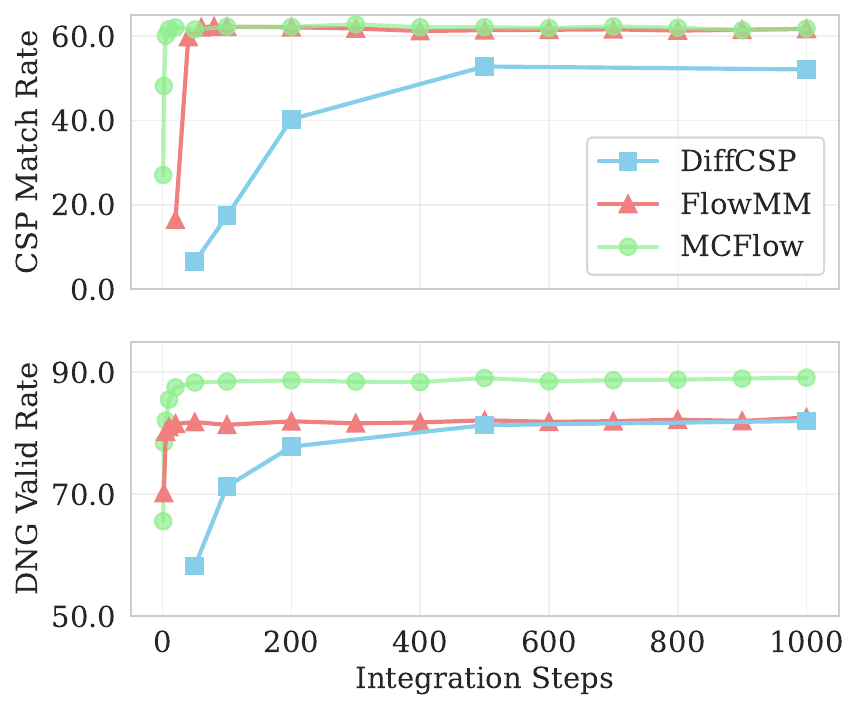} 
    \caption{\textbf{Effect of the number of integration steps on performance.} MCFlow reaches competitive crystal structure prediction (CSP) match rate and \emph{de novo} generation (DNG) validity with far fewer integration steps than FlowMM and DiffCSP.}
    \label{fig:integration_steps}
\end{figure}
\textbf{Results.} 
MCFlow achieves competitive performance on MP-20 and improved performance on MPTS-52, which contains larger unit cells (\cref{tab:csp}). The proposed ordering and augmentation drive these gains (\cref{tab:ablation,tab:ordering_ablation}); the larger margin on MPTS-52 reflects the DiT's scaling advantage on longer sequences. MCFlow saturates significantly earlier than FlowMM and DiffCSP in CSP match rate (\cref{fig:integration_steps}). We also observe consistent scaling behavior across model sizes, as detailed in \cref{fig:scaling}; on MP-20, multi-sample match rate saturates around the base model while RMSE continues to improve with scale, whereas MPTS-52 benefits from scale across both metrics. Noisy guidance (NG) further improves single-sample generation accuracy, as demonstrated by the ablation study in \cref{app:guidance}. Further qualitative samples illustrating structural diversity are provided in \cref{app:visualizations}. MCFlow also attains the highest polymorph-aware match rate (\cref{tab:metre}) and leads under both filtered and unfiltered match rate evaluations (\cref{tab:unfiltered_mr}).

\begin{figure*}[t]
    \centering
    \includegraphics[width=\textwidth]{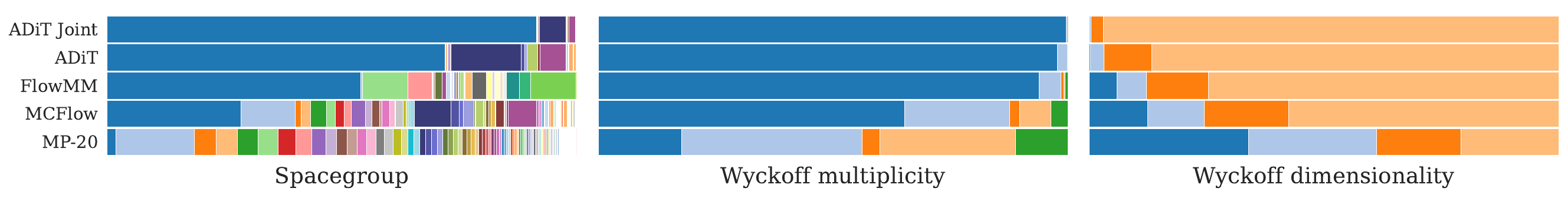}
    \vspace{-15pt}
    \caption{
    \textbf{Distributions of crystallographic symmetry properties.} MCFlow better captures crystallographic symmetry distributions than FlowMM and ADiT. From left to right, panels show the distributions of space groups
    (P1, Fm$\bar{3}$m, C2/m, P6$_3$/mmc, I4/mmm, Pnma, R$\bar{3}$m, Cmcm, Pm$\bar{3}$m, P4/mmm, P$\bar{1}$, $\ldots$),
    Wyckoff multiplicity (1, 2, 3, 4, 6),
    and Wyckoff dimensionality (0, 1, 2, 3).
    Space groups and Wyckoff positions are determined using the \texttt{SpaceGroupAnalyzer} module in \texttt{pymatgen} with distance of 0.1~\AA{}.
    }
    \label{fig:spacegroup}
\end{figure*}
\subsection{De Novo Generation}
In the DNG task, the model generates atom types, lattice parameters, and fractional coordinates at once.

\textbf{Metrics.} We evaluate DNG using the standardized LeMat-GenBench~\citep{lemat2025genbench} protocol, which provides a unified evaluation pipeline across models. All metrics are computed for 2{,}500 sampled crystals using an ensemble of three MLIPs---MACE-MP~\citep{batatia2024macemp}, UMA~\citep{wood2025uma}, Orb-v3~\citep{rhodes2025orbv3}---with LeMat-Bulk~\citep{siron2025lematbulk} as the reference set. We report validity, uniqueness, novelty, formation energy ($E_f$), energy above hull ($E_{\text{hull}}$), relaxation RMSD, stability ($E_{\text{hull}} \le 0$ eV/atom), metastability ($0 < E_{\text{hull}} \le 0.1$ eV/atom, disjoint from stability), S.U.N., and MS.U.N. rates (definitions in \cref{app:metrics}).

\textbf{Baselines.}
We compare MCFlow against representative crystal generative models on LeMat-GenBench, including
MatterGen~\citep{zeni2023mattergen},
PlaID++~\citep{xu2025plaidpp},
WyFormer~\citep{kazeev2025wyckoff},
DiffCSP~\citep{jiao2023crystal},
DiffCSP++~\citep{jiao2024space},
ADiT~\citep{joshi2025all},
LLaMat-2/3~\citep{mishra2025llamat},
SymmCD~\citep{levy2025symmcd},
CrystalFormer~\citep{crystalformer2025},
and Crystal-GFN~\citep{crystalgfn2025}.

\textbf{Results.}
MCFlow achieves strong performance on LeMat-GenBench across both evaluation groups (\cref{tab:lemat}). MCFlow leads on most metrics with and without MLIP pre-relaxation, particularly on stability-related ones (stability, metastability, $E_{\text{hull}}$), which we attribute to the proposed ordering combined with the DiT backbone. Under pre-relaxation, we further observe a stability--novelty trade-off across model scales: larger variants produce more stable but less novel samples, and MCFlow-S consequently leads MS.U.N.\ by retaining sample diversity.
Notably, MCFlow better captures crystallographic symmetry distributions over space groups and Wyckoff properties, whereas FlowMM and ADiT exhibit mode collapse toward lower-symmetry groups such as P1 (\cref{fig:spacegroup}).
These results indicate that MCFlow internalizes crystallographic symmetry from data without explicit symmetry constraints.

\begin{figure*}[t]
    \centering
    \includegraphics[width=\textwidth]{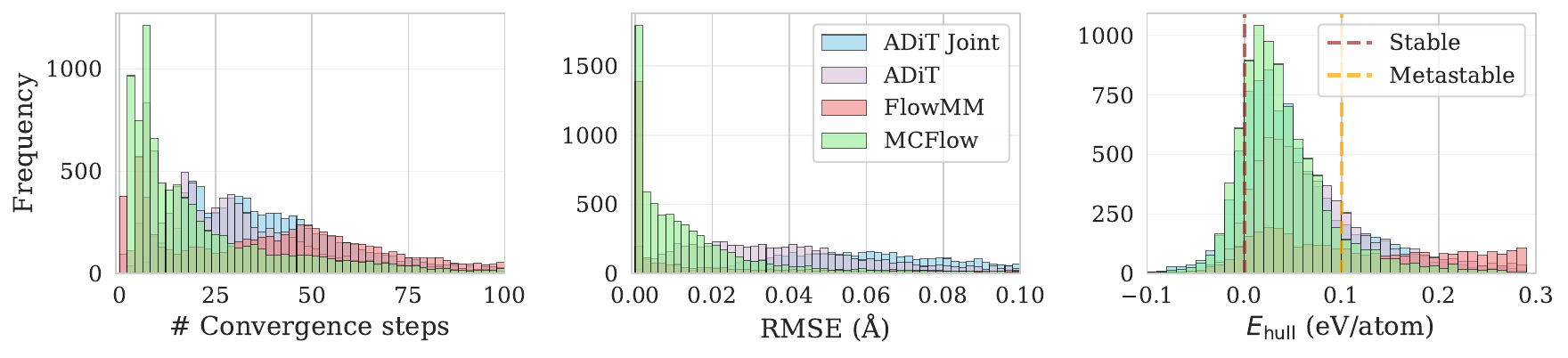}
    \vspace{-15pt}
    \caption{
    \textbf{Distributions of relaxed structures.}
    MCFlow samples are closer to local energy minima than FlowMM and ADiT, with smaller relaxation RMSE, fewer convergence steps, and lower $E_{\text{hull}}$.
    From left to right, plots show distributions of convergence steps, RMSE between initial and relaxed structures, and $E_{\text{hull}}$ along CHGNet~\citep{deng2023chgnet} geometry optimization trajectories.
    The convergence rates are 99.73/60.79/87.65/99.54\% for MCFlow/FlowMM/ADiT/ADiT Joint.
    }

    \label{fig:dng_stab}
\end{figure*}
MCFlow also produces geometrically stable samples. The distribution profiles in \cref{fig:dng_stab} demonstrate that MCFlow samples are closer to local energy minima as a population, exhibiting shorter relaxation trajectories and lower $E_{\text{hull}}$ values than the baselines. MCFlow achieves competitive performance with a relatively small number of integration steps (\cref{fig:integration_steps}). MCFlow maintains high structural and compositional validity on the larger MPTS-52 dataset (\cref{tab:dng_mpts52}).
While different DNG metrics exhibit varying trends across model sizes, validity shows consistent improvement, as detailed in \cref{fig:scaling}. Additional qualitative samples are provided in \cref{app:visualizations}.

\subsection{Atom Type Generation}
\begin{table}[t]
\centering
\caption{
\textbf{Atom type generation on MP-20 and MPTS-52.}
MCFlow shows higher compositional validity (Comp) and MR than the baselines when sampling multiple candidates per structure.
Comp follows \citet{xie2021crystal}; details in \cref{app:metrics}.}
\label{tab:atg}
\resizebox{\columnwidth}{!}{
\begin{tabular}{lcccc}
\toprule
\multirow{2}{*}{Model} & \multicolumn{2}{c}{MP-20} & \multicolumn{2}{c}{MPTS-52} \\
\cmidrule(lr){2-3} \cmidrule(lr){4-5}
& Comp (\%) $\uparrow$ & MR (\%) $\uparrow$ & Comp (\%) $\uparrow$ & MR (\%) $\uparrow$ \\
\midrule
CGCNN & 87.06 & 4.71 & 82.08 & 3.50 \\
FlowMM (1) & 83.39 & 2.40 & 76.12 & 0.88 \\
\midrule
MCFlow-S (1) & 88.97 & 1.57 &  83.14 & 1.01 \\
MCFlow-B (1)  & 90.17 & 3.12 & 83.56 & 1.20 \\
MCFlow-L (1)  & 90.07 & 3.32 & 83.67 & 1.24 \\
MCFlow-S (20) & 88.71 & \textbf{15.34} & 83.22 & \textbf{7.22} \\
MCFlow-B (20) & 90.22 & 14.73 & 84.17 & 5.72 \\
MCFlow-L (20) & \textbf{90.23} & 15.30 & \textbf{84.25} & 7.20 \\
\bottomrule
\end{tabular}
}
\end{table}
We consider the atom type generation task, abbreviated ATG, where we generate atom types conditioned on a crystal structure to demonstrate that the proposed multimodal flow
model naturally supports structure-to-composition generation using the same model.
This setting reflects a common case in materials datasets, where nearly identical crystal structures can support multiple chemically valid compositions. As a result, it is particularly relevant for exploring atomic substitution patterns within a fixed structure. To our knowledge, structure-conditioned ATG has not been studied as a stand-alone generative task in prior crystal generative modeling work; we therefore adapt the deterministic property-prediction model CGCNN~\citep{xie2018crystal} and the generative model FlowMM~\citep{miller2024flowmm} to the ATG task as reference points for compositional validity and accuracy.

\begin{table*}[t]
\centering
\caption{
\textbf{Atom type substitutions across four MP-20 test-set materials.}
MCFlow recovers chemically meaningful substitutions in different applications: the top-3 predicted elements mostly correspond to known materials sharing the same space group as the original structure.
For each case, we mask one element type and draw 100 samples using inference-time guidance built on atom type generation.
The four parent structures, all drawn from the MP-20 test set, are AcGaO$_3$ (mp-1183053, cubic Pm$\bar{3}$m perovskite oxide), Zr$_4$Fe$_4$Si$_4$ (mp-1102452, orthorhombic Pnma silicide intermetallic for hydrogen storage), Mn$_2$VGa (mp-21235, cubic Fm$\bar{3}$m full Heusler spintronic half-metal), and Ca$_2$In$_4$O$_8$ (mp-22766, cubic Fd$\bar{3}$m spinel photocatalyst).
A blank in the MP ID column means no matching Materials Project entry exists; a ``---'' in the Split column means the entry is not in MP-20.
}
\label{tab:masked_atg}
\resizebox{\textwidth}{!}{%
\begin{tabular}{clllllllllllllllll}
\toprule
\multirow{2}{*}{Rank} & \multicolumn{4}{c}{AcGaO$_3$ (mask Ac)} & \multicolumn{4}{c}{Zr$_4$Fe$_4$Si$_4$ (mask Fe)} & \multicolumn{4}{c}{Mn$_2$VGa (mask V)} & \multicolumn{4}{c}{Ca$_2$In$_4$O$_8$ (mask Ca)} \\
\cmidrule(lr){2-5} \cmidrule(lr){6-9} \cmidrule(lr){10-13} \cmidrule(lr){14-17}
 & Element & Freq (\%) & MP ID & Split & Element & Freq (\%) & MP ID & Split & Element & Freq (\%) & MP ID & Split & Element & Freq (\%) & MP ID & Split \\
\midrule
1 & Ac & 23 & 1183053 & Test  & Co & 40 & 1102565 & Test  & Mn & 23 & 999550  & ---   & Mn & 32 & 35162 & Train \\
2 & Yb & 22 & 1187561 & ---   & Cu & 32 & 22522   & Train & V  & 16 & 21235   & Test  & Ca & 19 & 22766 & Test  \\
3 & Sr & 10 &         & ---   & Ni & 19 & 1103272 & Val   & Nb & 15 & 1185972 & Train & Cd & 12 & 19803 & Val   \\
\bottomrule
\end{tabular}}
\end{table*}

\textbf{Results.} MCFlow generates compositionally valid and diverse atom types when conditioned on a fixed crystal structure. Compared to the baselines, MCFlow exhibits greater diversity across multiple samples per structure while maintaining higher compositional validity (\cref{tab:atg}). Further qualitative samples illustrating compositional diversity in atom type generation are provided in \cref{app:visualizations}. These results illustrate that a single fixed crystal structure can be compatible with multiple valid compositions.

\textbf{Atom type substitution.}
Building on atom type generation, MCFlow can perform atom type substitution: at each ATG step, we sample atom types from the CTMC posterior and overwrite the atom types at the known sites with given types, so only the masked sites are sampled.
Across four structurally and functionally diverse MP-20 test-set materials---a cubic Pm$\bar{3}$m perovskite oxide (AcGaO$_3$), an orthorhombic Pnma silicide intermetallic for hydrogen storage (Zr$_4$Fe$_4$Si$_4$), a cubic Fm$\bar{3}$m full Heusler spintronic half-metal (Mn$_2$VGa), and a cubic Fd$\bar{3}$m spinel photocatalyst (Ca$_2$In$_4$O$_8$)---masking one element and drawing 100 samples yields chemically meaningful top substitutions (\cref{tab:masked_atg}): the top-3 predicted elements mostly correspond to known Materials Project entries that share the same space group as the original structure.
This indicates that MCFlow has learned physically meaningful atom substitution patterns from data, without substitution rules.

\subsection{Ablation Studies}
We investigate the impact of composition- and symmetry-aware atom ordering and inter- and intra-orbit permutation augmentation to quantify component contributions. We evaluate these strategies using Match Rate and RMSE for CSP, and structural and compositional validity for DNG.

\textbf{Results.} Lexicographical ordering based on electronegativity and Wyckoff position provides compositional and crystallographic priors that are exploited by the DiT model, as shown in \cref{tab:ablation}. The PosEmb ablation removes explicit positional information from the sequence representation to assess the contribution of the proposed atom ordering. The resulting performance degradation indicates that the ordering encodes compositional and symmetry information (\cref{tab:ablation}). Inter- and intra-orbit permutation augmentation further improves performance by promoting robustness to symmetry-preserving permutations within this ordering. Furthermore, electronegativity-based ordering consistently outperforms atomic numbers, and the proposed hierarchical augmentation significantly outperforms full $N!$ random permutation (\cref{tab:ordering_ablation}). We also show that unified training matches CSP-only training (\cref{tab:unified_ablation}), and that decoupled time axes are critical for CSP: coupling $t{=}s$ substantially degrades CSP performance, and DiffCSP/FlowMM adapted to CSP fail to effectively condition on fixed compositions due to their coupled noise schedules (\cref{tab:coupled_ablation}).
\begin{table}[t]
\centering
\caption{
\textbf{Ablation results on MP-20 and MPTS-52.}
Removing the proposed hierarchical permutation augmentation and atom ordering leads to consistent performance degradation across tasks.
Rows denote cumulative ablations from the full MCFlow model
by removing inter-orbit permutation ($-\,$Inter), intra-orbit permutation ($-\,$Intra),
and positional embedding ($-\,$PosEmb).
MR and RMSE are reported for CSP, while Struct and Comp denote structural
and compositional validity for DNG; details in \cref{app:metrics}.
}
\label{tab:ablation}
\resizebox{\linewidth}{!}{%
\begin{tabular}{lcccccccc}
\toprule
\multirow{2}{*}{Model} & \multicolumn{4}{c}{MP-20} & \multicolumn{4}{c}{MPTS-52} \\
\cmidrule(lr){2-5} \cmidrule(lr){6-9}
& MR & RMSE & Struct & Comp
& MR & RMSE & Struct & Comp \\
\midrule
MCFlow     & 63.14 & 0.0611 & 99.44 & 89.85 & 26.46 & 0.1572 & 98.34 & 85.57 \\
\midrule
$-\,$Inter & 60.41 & 0.0772 & 99.48 & 89.64 & 22.80 & 0.1927 & 98.12 & 85.20 \\
$-\,$Intra & 58.54 & 0.1023 & 99.35 & 89.30 & 18.80 & 0.2231 & 97.70 & 84.83 \\
$-\,$PosEmb& 54.83 & 0.0979 & 99.48 & 85.30 & 16.44 & 0.2105 & 98.20 & 75.20 \\
\bottomrule
\end{tabular}
}
\end{table}

\section{Conclusion}

We introduced MCFlow, a unified multimodal flow framework that formulates crystal generation tasks as different inference trajectories through a single joint generative model. By decoupling flow time axes for atom types and crystal structures, MCFlow enables flexible conditional generation, allowing crystal structure prediction, \emph{de novo} generation, and atom type generation to be realized within a single model without task-specific architectures or retraining.

To instantiate this unified framework in an all-atom transformer architecture, we proposed a composition- and symmetry-aware atom ordering together with hierarchical permutation augmentation, providing a soft inductive bias that allows composition and crystallographic symmetry to be learned directly from data without explicit templates. Experiments demonstrate that a single MCFlow model achieves competitive performance across multiple crystal generation tasks, while reducing collapse toward low-symmetry space groups compared to existing baselines.

\textbf{Limitations.}
The composition- and symmetry-aware ordering exploits crystallographic symmetry and is therefore inapplicable to amorphous or disordered systems, which would require alternative inductive biases not relying on well-defined space groups. We further observe a stability--novelty trade-off across model scales (\cref{tab:lemat}): larger variants yield more stable but less novel samples, biasing generation toward known regions of materials space.

A natural next step is extending the any-to-any formulation to material property modalities, enabling property-conditioned generation within the same framework. A complementary direction is integrating MCFlow into agent-based, closed-loop materials discovery~\citep{takahara2025accelerated}, where its flexible generation modes can propose candidates under different conditioning settings.

\section*{Impact Statement}
MCFlow offers a computational tool for discovering next-generation materials, with applications including solid-state electrolytes, green-hydrogen catalysts, and advanced semiconductors. By enabling any-to-any modality generation in a single model, it supports diverse materials generation tasks without task-specific architectures.

\newpage
\bibliography{example_paper}
\bibliographystyle{icml2026}

\newpage
\appendix
\numberwithin{table}{section}
\numberwithin{figure}{section}
\onecolumn

\newpage
\section{Notations and Backgrounds}
\label{app:background}

\begin{table}[h]
\caption{Summary of Notations}
\label{tab:notation}
\begin{center}
\begin{small}
\begin{tabular}{ll}
\toprule
Symbol & Description \\
\midrule
$C$ & Infinite periodic crystal \\
$N$ & Number of atoms in the primitive unit cell \\
$A$ & Atom types sequence ($[a_i]_{i=1}^N \in \mathcal{A}^{1 \times N}$) \\
$F$ & Fractional coordinates ($[f_i]_{i=1}^N \in [0, 1)^{3 \times N}$) \\
$X$ & Cartesian coordinates ($X=LF$) \\
$L$ & Lattice matrix \\
$L^l$ & Lattice lengths ($\{a, b, c\}\in\mathbb{R}^{1\times 3}$)\\
$L^a$ & Lattice angles ($\{\alpha, \beta, \gamma\}\in \left[60^\circ,120^\circ\right]^{1\times3}$) \\
$G$ & Space group \\
$O$ & Orbits ($\{g\cdot f\mid g\in G\}$) \\
$G_{f}$ & Site symmetry group ($\{g\mid g\cdot f=f\}$) \\
$\mathcal{W}$ & Wyckoff position \\
\midrule
$\mathcal{S}$ & Discrete state space (e.g., atom type space $\mathcal{A}$) \\
$u(x_t, t)$ & Velocity field (continuous) or rate vector (discrete) \\
$p_t$ & Probability path at time $t$ \\
$t, s$ & Flow time coordinates ($t$: atom types, $s$: structure) \\
$\pi$ & Coupling distribution between $p_0$ and $p_1$ \\
$\delta_{x_t}$ & Probability vector with $\delta_{x_t}(x_t)=1$ on discrete space\\ 
\midrule
$\mathbb{T}^{3\times N}$ & Torus manifold for fractional coordinates ($\mathbb{R}^{3\times N} / \mathbb{Z}^{3\times N}$) \\
$\text{proj}(\cdot)$ & Projection onto the torus ($x \pmod 1$) \\
\bottomrule
\end{tabular}
\end{small}
\end{center}
\end{table}

We first formalize periodic crystals and symmetry (space groups, Wyckoff positions), then review flow-based generative models on various state spaces. \cref{tab:notation} summarizes the key notations used throughout this paper.

\subsection{Crystal Representation}
We represent crystals as infinite periodic arrangements of primitive unit cells. The unit cell is represented by atom types $A = [a_i]_{i=1}^{N} \in \mathbb{Z}^{1 \times N}$, fractional coordinates $F = [f_i]_{i=1}^{N} \in [0, 1)^{3 \times N}$, and a lattice matrix $L=[l_i]_{i=1}^{3} \in \mathbb{R}^{3 \times 3}$. While there are many possible choices of primitive unit cells, we use Niggli reduction~\citep{grosse2004numerically} to fix a canonical representative (up to numerical tolerance). The lattice matrix $L$ can be also represented as roto-translation invariant lattice parameters: three side lengths $L^{l}=\{a, b, c\}\in\mathbb{R}^{1\times 3}$ and three internal angles $L^{a}=\{\alpha, \beta, \gamma\}\in \left[60^\circ,120^\circ\right]^{1\times3}$ of the unit cell. Given this lattice representation $L$, the 3D Cartesian coordinates $X=[x_i]_{i=1}^{N} \in \mathbb{R}^{3\times N}$ are obtained by the linear combination of lattice vectors with fractional coefficients: $X=LF$. The infinite periodic crystal $C$ is then defined as:
\begin{equation}
    C = \{ (a_i, x_i + Lk) \mid i \in \{1, \dots, N\},\quad k \in \mathbb{Z}^{3} \}
\end{equation}
where $k_j$ denotes the integral translation along the $j$-th lattice vector $l_{j}$.

\subsection{Crystal Symmetry}
Space groups~\citep{hiller1986crystallography} characterize the invariances of infinite periodic structures.

\textbf{Space groups.}
A space group $G$ is a group of isometries that tile $\mathbb{R}^3$ with repeating unit cells, i.e., a set of operation $g$ under which the transformed crystal $C$ is identical to the original crystal
where operation $g=(R, t) \in G$ consists of an orthogonal matrix $R \in O(3)$ and a translation $t \in \mathbb{R}^3$.
Its action on a Cartesian coordinate $x \in \mathbb{R}^3$ denotes $g \cdot x \coloneqq Rx + t$.
When expressed in fractional coordinates $f$ with $x=Lf$, we use the corresponding induced action
\begin{equation}
    g \cdot f \coloneqq L^{-1}(R(Lf)+t)\pmod 1,  
\end{equation}
where $\pmod 1$ is applied component-wise, identifying fractional coordinates that differ by lattice translations.
A space group operation first acts on atomic positions in Cartesian space and then maps the result back into the unit cell via fractional coordinates modulo lattice translations.

A crystal $C$ is said to be invariant under space group $G$ if for every $g \in G$, there exists a permutation $\sigma_g \in S_N$ of the atom indices such that the transformed atomic positions coincide with the original ones up to lattice translations and a permutation of atom indices preserving atom types. This permutation $\sigma_g$ rearranges indices changed by the symmetry operation $g$ while preserving their atom types. Every crystal has space group symmetry which is one of the 230 distinct space groups, a unique way a crystal's atoms can be arranged periodically.

\textbf{Wyckoff positions.}
Given a space group $G$, Wyckoff positions~\citep{wyckoff1930analytical} describe crystal symmetry at the level of atomic sites. For a fractional coordinate $f$, a set of equivalent atoms generated by group $G$, called \textit{orbit}, is
\[
G\cdot f \coloneqq \{ g \cdot f \mid g \in G \},
\]
\textit{Site symmetry group} of $f$ is a subgroup of the space group consisting of operations that leave the site invariant (up to lattice translations):
\[
G_{f}\coloneqq \{ g \in G \mid g \cdot f = f \}.
\]
We can classify site symmetry groups based on conjugate relation, which is equivalent relation, up to basis change by space group, between $G_f$ and $G_{f'}$ defined by 
$G_{f}=gG_{f'}g^{-1}$ for some $g\in G$.
The conjugate relation induces an equivalence class $[G_f]$ of site symmetry groups
\[
[G_f] \coloneqq \{ g G_f g^{-1} \mid g \in G \}.
\]
\textit{Wyckoff position} is a set of orbits whose site symmetry groups have conjugate relation:
\[
\mathcal{W}_{[G_{f}]}\coloneqq\{G\cdot f' \mid G_{f'}\in[G_{f}]\}.
\]
The size of its orbit $|G \cdot f|$ is called the \textit{Wyckoff multiplicity}. Wyckoff positions are conventionally labeled by letters $a,b,c,\ldots$, called the \textit{Wyckoff letter}. The letter is assigned according to the site symmetry following the IUCr convention~\citep{prince2004international}. Ordering atoms by Wyckoff position encodes crystallographic symmetry
information that is useful for transformer models.

\subsection{Flow Generative Model}
We provide a self-contained background on flow-based models and flow matching in various state spaces. For a more comprehensive guide and practical implementation details of flow matching, we refer the reader to \citet{lipman2024flowmatchingguidecode}. Flow models naturally extend to multimodal settings by combining
continuous and discrete state spaces under a unified generator matching framework~\citep{holderrieth2024generator}.

\textbf{Flow-based generative model.}
Flow-based generative model~\citep{chen2018neural}, or simply flow model, transports a base distribution
$p_0(x_0)=p_{\mathrm{base}}$
to data distribution
$p_1(x_1)=p_{\mathrm{data}}$
by solving an ordinary differential equation (ODE)
characterized by a time-dependent velocity field
$u(x_t,t)$ for $t\in[0,1]$:
\begin{equation}
\frac{d x_t}{dt} = u(x_t,t), \quad x_0 \sim p_0.
\end{equation}
Flow models construct and learn $u(x_t,t)$ which corresponds to probability path $\{p_t\}_{t\in[0,1]}$ evolving from $p_0=p_{\mathrm{base}}$ to $p_1=p_{\mathrm{data}}$.

\textbf{Flow matching.}
Flow matching (FM) is a simulation-free training method to learn the constructed
velocity field $u(x_t,t)$~\citep{lipman2022flow,liu2022rectifiedflow}.
Given a neural network $u_\theta(x_t,t)$,
the objective is to minimize the squared distance
\begin{equation} \label{eq:fm}
\mathcal{L}_{\text{FM}}(\theta)=\mathbb{E}_{t, x_t}
\left[
\lVert u_\theta(x_t,t) - u(x_t,t) \rVert^2
\right],
\end{equation}
where $t\sim \mathcal{U}[0, 1]$ and $x_{t}\sim p_t$. \citet{lipman2022flow} defines the probability path as a mixture of tractable conditional probability paths $p_t(\cdot|x_0, x_1)$:
\begin{equation}
    p_t(x) = \int p_t(x|x_0, x_1) d\pi(x_0, x_1),
\end{equation} 
where $\pi(x_0, x_1)$ is a independent coupling distribution over the base and data distributions, i.e., $\pi(x_0, x_1)=p_{0}(x_0)p_{1}(x_1)$. \citet{lipman2022flow} introduces conditional flow matching (CFM) which minimizes the distance between $u_{\theta}(x_t,t)$ with tractable conditional velocity field $u(x_t,t|x_0,x_1)$:
\begin{equation}
\mathcal{L}_{\text{CFM}}(\theta)=\mathbb{E}_{t, x_0, x_1, x_t}
\left[
\lVert u_\theta(x_t,t) - u(x_t,t|x_0,x_1) \rVert^2
\right],
\end{equation}
where pairs $(x_0, x_1)$ are sampled from the coupling distribution $\pi(x_0, x_1)$, and conditional flow $x_t$ is sampled from conditional probability path $p_t(\cdot|x_0, x_1)$. The conditional flow matching objective induces the same gradient with respect to $\theta$ as the marginal flow matching objective, i.e., $\nabla_\theta \mathcal{L}_{\text{FM}}(\theta)
=\nabla_\theta \mathcal{L}_{\text{CFM}}(\theta)$

A simple and commonly used conditional flow is linear interpolation $x_t = (1-t)x_0 + t x_1$ between two endpoints $x_0$ and $x_1$, corresponding to the conditional velocity field:
\begin{equation}
u(x_t,t \,|\, x_0,x_1) = \frac{x_1-x_t}{1-t}.
\end{equation}
We can construct the marginal velocity field $u(x_t, t)$, constructed via posterior expectation over $\pi$:
\begin{equation} \label{eq:posterior_vel}
u(x_t,t)=\mathbb{E}_{x_0, x_1}
\big[u(x_t,t|x_0,x_1)\,|\, x_t\big],
\end{equation}
ensuring transports from $p_0$ to $p_1$. The CFM objective extends the squared distance loss to a Bregman
divergence $D_\Phi(u,v)=\Phi(u)-\Phi(v)-\langle\nabla\Phi(v),u-v\rangle$:
\begin{equation}\label{eq:cfm}
\mathbb{E}_{t, x_0, x_1, x_t}
\left[D(u_\theta(x_t,t),u(x_t,t|x_0,x_1))\right],
\end{equation}
where $\Phi$ is a convex function. Bregman divergences are used since they
preserve the gradient equivalence between conditional and marginal flow
matching objectives. This formulation allows different choices of
$\Phi$ and recovers the squared distance loss $D(u,v)=\lVert u-v \rVert^{2}$ when
$\Phi(u)=\|u\|^2$.

\textbf{Riemannian flow model.}
The above flow model assumes that data lie in Euclidean space.
When data distribution is on Riemannian manifold $\mathcal{M}$, Riemannian flow model~\citep{chen2023riemannian} transports base distribution on the manifold to the data distribution with velocity field $u(x_t,t)$ in tangent space $T_{x_t}\mathcal{M}$. The construction $u(x_t,t)$ from the conditional velocity field $u(x_t,t \,|\, x_0,x_1)$ is obtained by replacing straight-line paths with geodesics: $x_t = \exp_{x_0}(t\log_{x_0}(x_1))$ where $\exp_x$ and $\log_x$ denote the Riemannian exponential and logarithmic maps. The corresponding conditional velocity field is
\begin{equation}
u(x_t,t|x_0,x_1)
= \frac{\log_{x_t}(x_1)}{1-t}
\in T_{x_t}\mathcal{M}.
\end{equation}
Flow matching regresses $u_\theta(x_t,t)$ onto the target $u(x_t,t|x_0,x_1)$ in the tangent space. A commonly used Bregman divergence is $D(u,v)=\lVert u-v \rVert_{g}^{2}$ for a Riemannian metric $g$ on the manifold $\mathcal{M}$.

\textbf{Discrete flow model.}
To extend flow models on continuous space to flow model on discrete state space $\mathcal{S}$, we convert solving ODE with the velocity field $u(x_t,t)$ into transition probability vector $p_{t+h|t}(\cdot \,|\, x_t)$ from current state $x_t$ for small increment $h>0$, called continuous-time Markov chain (CTMC):
\begin{equation}
p_{t+h|t}(\cdot \,|\, x_t)
= \delta_{x_t} + h\,u(x_t,t) + o(h),
\end{equation}
where rate vector $u(x_t,t)$ governs CTMC and $\delta_{x_t}$ is the probability vector for current state $x_t$ which satisfies $\delta_{x_t}(x_t)=1$ and $\delta_{x_t}(x)=0$ if $x\neq x_t$. Discrete flow model aims to learn the rate vector $u(x_t,t)$ which transports the base probability vector $p_0$ to the data probability vector $p_1$ through probability vector path $\{p_{t}\}_{t\in[0,1]}$. We use masked flow model where the base probability vector $p_{0}=\delta_{M}$ for the masked state $M$.

\textbf{Discrete flow matching.}
Analogous to the continuous case, we match the parameterized and target rate vectors, $u_\theta(x_t, t)$ and $u(x_t, t)$, respectively, given the current state $x_t$. A commonly used Bregman divergence is the generalized KL divergence $D(u,v)=\sum_j u_j \log(u_j/v_j) - \sum_j u_j + \sum_j v_j$ and the FM objective has the same form with \cref{eq:cfm} where marginal rate vector are constructed by  conditional rate vector as in \cref{eq:posterior_vel}. Discrete flow model also extends to CFM objective for tractability. A commonly used conditional probability vector path is the independent mixture path between the base state $x_0=M$ and the target state $x_1$ sampled from $p_1$:
\begin{equation}
p_t(\cdot\,|\, x_0,x_1) = (1-t)\delta_{x_0} + t\delta_{x_1},
\end{equation}
which is the discrete analog of linear interpolation. The corresponding conditional rate vector is given by:
\begin{equation}
u(x_t,t| x_0,x_1) = \frac{\delta_{x_0}(x_t)\delta_{x_1}}{1-t}.
\end{equation}
where $x_0=M$. While continuous and discrete flow matching are defined on different frameworks, ODE and CTMC, they can be unified into generator matching on continuous-time Markov process~\citep{holderrieth2024generator}. This unifying perspective enables joint generative processes in which
continuous and discrete dynamics evolve simultaneously over different components of the state space.
\newpage
\section{Multimodal Flow Details}
\label{app:mcflow}

\begin{table*}[t]
    \centering
\caption{
    \textbf{Configurations for Multimodal Flow Matching.}
    We leverage the flexibility of the flow matching framework to design components for each modality. 
    The Base dist. column specifies the prior distribution $p_0$, selected to match the support of each modality. 
    The Cond. target column defines the conditional target vector fields or rate vectors derived from data pairs, utilizing linear interpolation for Euclidean space, geodesic paths for the torus, and independent mixture paths for discrete atom types. 
    The Parameterization column describes how to approximate the target velocity field from the clean data prediction ($F_{\theta}, L^l_\theta, L^a_\theta$) or the target categorical posterior distribution via atom type logit prediction $A_{\theta}$. 
    Finally, the Bregman div. column lists the specific divergence metric minimized during training to match the parameterized and target velocity field or rate vector. Since the torus is flat and locally Euclidean, the Bregman divergence on fractional coordinates is the squared Euclidean distance in the tangent space.
    }
    \label{tab:modality_summary}
    \resizebox{\textwidth}{!}{
    \begin{tabular}{lcccc}
        \toprule
        Modality ($m$) &
        Base dist. $p_0$ &
        Cond. target &
        Parameterization $u^m_\theta$ &
        Bregman div. $D_{m}(u,v)$ \\
        \midrule
        \addlinespace[0.6em]
        Atom type ($A$) &
        $\delta_{M}$ &
        $\displaystyle \frac{\delta_{A_0}(A_t)\delta_{A_1}}{1-t}$ &
        $\displaystyle \frac{\delta_{A_0}(A_t)\text{Cat}(\text{softmax}(A_\theta))}{1-t}$ &
        $\displaystyle \sum_j u_j \log\!\left(\frac{u_j}{v_j}\right) - \sum_j u_j + \sum_j v_j$ \\
        \addlinespace[0.6em]
        Frac. coord. ($F$) &
        $\mathcal{U}[0,1)^{3\times N}$ &
        $\displaystyle \frac{\log_{F_s}(F_1)}{1-s}$ &
        $\displaystyle \frac{\log_{F_s}\left(\text{proj}(F_\theta)\right)}{1-s}$ &
        $\|u-v\|^2$ \\
        \addlinespace[0.6em]
        Length ($L^l$) &
        $\text{LogNorm}(\mu,\sigma)$ &
        $\displaystyle \frac{L^l_1 - L^l_s}{1-s}$ &
        $\displaystyle \frac{L^l_\theta - L^l_s}{1-s}$ &
        $\|u-v\|^2$ \\
        \addlinespace[0.6em]
        Angle ($L^a$) &
        $\displaystyle \mathcal{U}\left[60^{\circ},120^{\circ}\right]^{1\times3}$ &
        $\displaystyle \frac{L^a_1 - L^a_s}{1-s}$ &
        $\displaystyle \frac{L^a_\theta - L^a_s}{1-s}$ &
        $\|u-v\|^2$ \\
        \addlinespace[0.2em]
        \bottomrule
    \end{tabular}
    }
\end{table*}

We provide detailed configurations for each modality in \cref{tab:modality_summary}.

\textbf{Riemannian flow on the torus for fractional coordinates.}
We model fractional coordinates $F$ on the torus manifold
$\mathbb{T}^{3\times N}$ to enforce the periodicity of crystal
structures. The corresponding Riemannian exponential and logarithmic
maps are given by
\begin{equation}
    \exp_{f}(v) = (f+v) \pmod 1, \quad
    \log_f(f') = \frac{1}{2\pi}\mathrm{atan2}\!\left(
    \sin(2\pi(f'-f)), \cos(2\pi(f'-f))\right).
\end{equation}
This Riemannian formulation ensures that generated fractional
coordinates remain within the fundamental domain $[0,1)$ by construction.

\textbf{Parameterization details.}
We use $x_1$-prediction across all modalities and convert the predicted
clean data into velocity fields or rate vectors according to
the underlying state space of each modality. For lattice lengths and angles in Euclidean space, the network outputs
$L_{\theta}^{l}$ and $L_{\theta}^{a}$, which are converted to conditional
velocity targets via
$(L_{\theta}^{l}-L^{l}_{s})/(1-s)$ and
$(L_{\theta}^{a}-L^{a}_{s})/(1-s)$, respectively. For fractional coordinates on the torus $\mathbb{T}^{3\times N}$, the
network predicts $F_{\theta}$, which is first projected component-wise
onto the torus via $\mathrm{proj}(F)=F \pmod 1$. The velocity field is
then obtained by applying the logarithmic map on the torus,
$\log_{F_s}\!\left(\mathrm{proj}(F_\theta)\right)/(1-s)$, ensuring that
the resulting flow respects the periodic topology of fractional
coordinates. For discrete atom types, the network predicts logits $A_\theta$
corresponding to the target atom types $A_1$. These logits define a
posterior categorical distribution
$p_\theta=\mathrm{Cat}(\mathrm{softmax}(A_\theta))$, which induces the rate vector of the CTMC as
$\delta_{A_0}(A_t)p_\theta/(1-t)$. To mitigate numerical instability near the terminal times, we clip the
flow times $t$ and $s$ to $0.9$ during training, following
\citet{campbell2024generative}.

\textbf{Base distribution details.}
For lattice lengths, the parameters of the LogNormal prior are fitted to training data statistics via maximum likelihood. During inference, the number of atoms $N$ is sampled from the empirical distribution of the training set, which determines the dimensionality of the base distribution.

\newpage
\section{Algorithm}
\label{app:algorithm}

\begin{algorithm}[ht]
\caption{MCFlow preprocessing and training}
\label{alg:training}
\begin{algorithmic}[1]
\STATE \textbf{Input:} dataset $\mathcal{D}$, clean data (and logit) prediction network $f_{\theta}$, joint base dist. $p_{0}$, weights $\{w_m\}_{m\in \{\text{A},\text{F},\text{L}^{a},\text{L}^{l}\}}$.

\STATE \textbf{Preprocessing:} 
\STATE Group atoms into orbits for each crystal in $\mathcal{D}$.
\STATE Sort orbits lexicographically by electronegativity and Wyckoff position for each crystal.

\textbf{Training:} 
\STATE Sample $(A_1,X_1)\sim\mathcal{D}$ where $X_1=(F_1,L^l_1,L^a_1)$.
\STATE Apply inter- and intra-orbit permutations. 
\STATE Apply global modulo translation.
\STATE Sample $t,s\sim \mathcal{U}[0,1]$ and $(A_0,F_0,L^l_0,L^a_0)\sim p_{0}(A_0,X_0)$.
\STATE Sample (or evaluate) intermediates from endpoint-conditioned paths:
\[
A_t\sim\mathrm{Cat}((1-t)\delta_{A_0} + t\delta_{A_1}), \quad F_s\leftarrow\exp_{F_0}(s\log_{F_0}(F_1)),\quad L^l_s\leftarrow(1-s)L^l_0+sL^l_1,\quad
L^a_s\leftarrow(1-s)L^a_0+sL^a_1.
\]
\STATE Construct modality-wise conditional target velocity field or rate vector:
\[
u^{\text{A}}=\frac{\delta_{A_0}(A_t)\delta_{A_1}}{1-t},\quad u^{\text{F}}\leftarrow\frac{\log_{F_s}(F_1)}{1-s},\quad
u^{\text{L}^{l}}\leftarrow \frac{L^l_1 - L^l_s}{1-s},\quad u^{\text{L}^{a}}\leftarrow \frac{L^a_1 - L^a_s}{1-s}.
\]
\STATE Predict $\left(A_{\theta}, F_{\theta},L^{l}_{\theta},L^{a}_{\theta}\right) \leftarrow f_{\theta}(A_t,X_s,t,s)$.

\STATE Compute rate vector for atom types and velocity fields for crystal structure:
\[
u^{\text{A}}_{\theta}\leftarrow \frac{\delta_{A_0}(A_t)\text{Cat}(\text{softmax}(A_\theta))}{1-t},\quad u^{\text{F}}_{\theta}\leftarrow \frac{\log_{F_s}\left(\text{proj}(F_\theta)\right)}{1-s},\quad u^{\text{L}^{l}}_{\theta}\leftarrow \frac{L^l_\theta - L^l_s}{1-s},\quad u^{\text{L}^{a}}_{\theta}\leftarrow \frac{L^a_\theta - L^a_s}{1-s}
\]

\STATE Compute the multimodal flow matching objective:
\[
\widehat{\mathcal{L}}(\theta)=\sum_{m\in\{{\text{A},\text{F},\text{L}^{a},\text{L}^{l}\}}} w_m\,D_m\!\big(u^m_\theta,\,u^m\big),
\]

\STATE Update parameters $\theta$ using AdamW with learning rate $\eta$.
\end{algorithmic}
\end{algorithm}

\begin{algorithm}[t]
\caption{MCFlow any-to-any modality generation (DNG/CSP/ATG) during inference}
\label{alg:inference}
\begin{algorithmic}[1]
\STATE \textbf{Input:} trained model $\text{MCFlow}_{\theta}$, timesteps $K$, empirical atom-count distribution $p_{\text{emp}}(N)$.
\STATE \textbf{Input:} task $\tau\in\{\text{DNG},\text{CSP},\text{ATG}\}$ with atom types $A_1$ (CSP); structure $X_1$ (ATG).
\STATE \textbf{Output:} a sample $(A,X)$.

\vspace{0.5em}
\STATE Set task-specific trajectory $\gamma$ and conditioning set $\mathcal{C}$.
\STATE \textbf{if} $\tau=\text{DNG}$: $\gamma(\lambda)=(\lambda,\lambda)$,\; $\mathcal{C}\leftarrow\emptyset$.
\STATE \textbf{if} $\tau=\text{CSP}$: $\gamma(\lambda)=(1,\lambda)$,\; $\mathcal{C}\leftarrow\{A\}$.
\STATE \textbf{if} $\tau=\text{ATG}$: $\gamma(\lambda)=(\lambda,1)$,\; $\mathcal{C}\leftarrow\{X\}$.

\vspace{0.5em}
\STATE Fix observed modality: $A\leftarrow A_1$ if $\tau=\text{CSP}$; $X\leftarrow X_1$ if $\tau=\text{ATG}$.
\STATE Set $N \leftarrow$ sample $p_{\text{emp}}(N)$ if $\tau=\text{DNG}$; otherwise $N=\lvert A_1\rvert$ (CSP) or $N=\lvert X_1\rvert$ (ATG).
\STATE If $A \notin \mathcal{C}$, sample $A \sim p_0(A)$.
\STATE If $X \notin \mathcal{C}$, sample $X \sim p_0(X)$.

\vspace{0.5em}
\FOR{$k=0$ \textbf{to} $K-1$}
    \STATE $(t_k,s_k)\leftarrow \gamma(k/K)$;\quad $(t_{k+1},s_{k+1})\leftarrow \gamma((k+1)/K)$
    \STATE $\Delta t \leftarrow t_{k+1}-t_k$;\quad $\Delta s \leftarrow s_{k+1}-s_k$
    \STATE $(u^{\text{A}}_\theta,u^{\text{X}}_\theta)\leftarrow \text{MCFlow}_{\theta}(A,X,t_k,s_k)$
    \STATE If $A \notin \mathcal{C}$, $A \leftarrow \text{CTMCStep}(A, u^{\text{A}}_\theta, \Delta t)$.
    \STATE If $X \notin \mathcal{C}$, $X \leftarrow \text{ODEStep}(X, u^{\text{X}}_\theta, \Delta s)$.
\ENDFOR
\STATE \textbf{return} $(A,X)$.
\end{algorithmic}
\end{algorithm}

\begin{algorithm}[t]
\caption{MCFlow inference with noisy guidance (CSP/ATG)}
\label{alg:ng}
\begin{algorithmic}[1]
\STATE \textbf{Input:} timesteps $K$, guidance scale $\omega$, noise level $\sigma$.
\STATE \textbf{Input:} task $\tau\in\{\text{CSP},\text{ATG}\}$ with atom types $A_1$ (CSP); structure $X_1$ (ATG).
\STATE \textbf{Output:} a guided sample $(A,X)$.

\vspace{0.5em}
\STATE Set task-specific trajectory $\gamma$ and conditioning set $\mathcal{C}$.
\STATE \textbf{if} $\tau=\text{CSP}$: $\gamma(\lambda)=(1,\lambda)$,\; $\mathcal{C}\leftarrow\{A\}$.
\STATE \textbf{if} $\tau=\text{ATG}$: $\gamma(\lambda)=(\lambda,1)$,\; $\mathcal{C}\leftarrow\{X\}$.

\vspace{0.5em}
\STATE Fix observed modality: $A\leftarrow A_1$ if $\tau=\text{CSP}$; $X\leftarrow X_1$ if $\tau=\text{ATG}$.
\STATE Set $N \leftarrow \lvert A_1\rvert$ (CSP) or $N \leftarrow \lvert X_1\rvert$ (ATG).

\vspace{0.5em}
\STATE \textbf{if} $\tau=\text{CSP}$: sample $X \sim p_0(X)$.
\STATE \textbf{if} $\tau=\text{ATG}$: sample $A \sim p_0(A)$.

\vspace{0.5em}
\FOR{$k=0$ \textbf{to} $K-1$}
    \STATE $(t_k,s_k)\leftarrow \gamma(k/K)$;\quad $(t_{k+1},s_{k+1})\leftarrow \gamma((k+1)/K)$
    \STATE $\Delta t \leftarrow t_{k+1}-t_k$;\quad $\Delta s \leftarrow s_{k+1}-s_k$

    \STATE \textbf{if} $\tau=\text{CSP}$:
    \STATE \quad Sample $A_0 \sim p_0(A)$ and $A_\sigma \sim \mathrm{Cat}((1-\sigma)\delta_{A_0}+\sigma\delta_{A_1})$
    \STATE \quad$u^X_{\text{cond}} \leftarrow u^X_\theta(A_1, X, 1, s_k)$;\quad
           $u^X_{\text{corr}} \leftarrow u^X_\theta(A_\sigma, X, \sigma, s_k)$
    \STATE \quad$u^X_{\text{ng}} \leftarrow (1-\omega)\,u^X_{\text{corr}} + \omega\,u^X_{\text{cond}}$
    \STATE \quad$X \leftarrow \text{ODEStep}(X, u^X_{\text{ng}}, \Delta s)$

    \STATE \textbf{if} $\tau=\text{ATG}$:
    \STATE \quad Sample $X_0 \sim p_0(X)$ and $X_\sigma \sim p_\sigma(X_\sigma \mid X_0,X_1)$
    \STATE \quad$u^A_{\text{cond}} \leftarrow u^A_\theta(A, X_1, t_k, 1)$;\quad
           $u^A_{\text{corr}} \leftarrow u^A_\theta(A, X_\sigma, t_k, \sigma)$
    \STATE \quad$u^A_{\text{ng}} \leftarrow (1-\omega)\,u^A_{\text{corr}} + \omega\,u^A_{\text{cond}}$
    \STATE \quad$A \leftarrow \text{CTMCStep}(A, u^A_{\text{ng}}, \Delta t)$
\ENDFOR
\STATE \textbf{return} $(A,X)$.
\end{algorithmic}
\end{algorithm}

\clearpage
\section{Ordering and Augmentation Analysis}
\label{app:ordering}

\begin{wrapfigure}{r}{0.4\textwidth}
    \centering
    \vspace{-10pt}
    \includegraphics[width=0.4\textwidth]{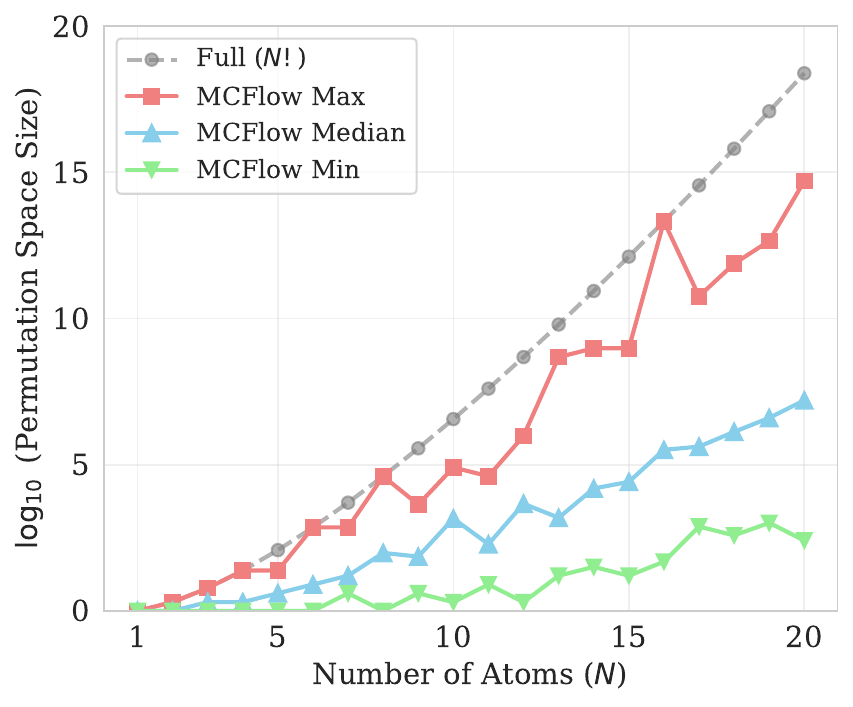}
    \vspace{-10pt}
    \caption{
    \textbf{Permutation space reduction.} 
    Logarithmic scale comparison of full permutation space $N!$ and reduced space $\prod_i |\mathcal{W}^i|! \prod_j |O^i_j|!$ across unit cell sizes $N=1$ to $20$.
    }
    \label{fig:permutation_reduction}
    \vspace{-10pt} 
\end{wrapfigure}
\textbf{Permutation space reduction analysis.}
To quantify the impact of our hierarchical permutation augmentation, we evaluate search space complexity in the MP-20 training set. For each crystal structure, we calculate the size of the reduced permutation space using the formula $\prod_i |\mathcal{W}^i|! \prod_j |O^i_j|!$, where $i$ denotes groups sharing the same element and Wyckoff position, and $j$ denotes individual orbits within those groups. We then group these values by the number of atoms in the unit cell ($N$) and calculate the minimum, median, and maximum values for each group to capture the distribution across different symmetry.

\cref{fig:permutation_reduction} illustrates the size of the permutation space as a function of $N$. While the naive permutation space grows factorially as $N!$, our proposed ordering combined with the hierarchical augmentation constraints drastically restricts the residual freedom. As shown in the figure, for a unit cell with $N=20$ atoms, the median reduced space size is approximately $10^{7.2}$, whereas the full permutation space is $20! \approx 10^{18.4}$. This reduction of over 11 orders of magnitude provides a strong inductive bias, enabling the Diffusion Transformer to effectively learn crystallographic symmetry and compositional priors directly from data without explicit structural templates.
\newpage
\section{Noisy Guidance}
\label{app:guidance}

\paragraph{Noisy guidance for atom type generation.}
For atom type generation, we apply noisy guidance at the logit level of the categorical distribution, which induces guided rate vector for the corresponding CTMC.

\begin{table}[t]
\centering
\caption{\textbf{Ablation study of noisy guidance on CSP and atom type generation (ATG).}
Noisy guidance improves structural fidelity in CSP and yields competitive performance in ATG across datasets.
Single-sample generation per composition or structure.
\textbf{Bold} indicates the best performance within each model size group.}
\label{tab:ng_ablation}
\resizebox{\textwidth}{!}{
\begin{tabular}{llcccc|cccc}
\toprule
\multirow{3}{*}{Model} & \multirow{3}{*}{NG} & \multicolumn{4}{c|}{CSP} & \multicolumn{4}{c}{ATG} \\
\cmidrule(lr){3-6} \cmidrule(lr){7-10}
& & \multicolumn{2}{c}{MP-20} & \multicolumn{2}{c|}{MPTS-52} & \multicolumn{2}{c}{MP-20} & \multicolumn{2}{c}{MPTS-52} \\
\cmidrule(lr){3-4} \cmidrule(lr){5-6} \cmidrule(lr){7-8} \cmidrule(lr){9-10}
& & MR (\%) $\uparrow$ & RMSE $\downarrow$ & MR (\%) $\uparrow$ & RMSE $\downarrow$ & Comp (\%) $\uparrow$ & MR (\%) $\uparrow$ & Comp (\%) $\uparrow$ & MR (\%) $\uparrow$ \\
\midrule
MCFlow-S & \redx & 54.40 & 0.1339 & 22.58 & 0.1880 & 88.89 & 1.53 & \textbf{83.19} & 0.98 \\
MCFlow-S & \greencheck & \textbf{56.10} & \textbf{0.1269} & \textbf{22.88} & \textbf{0.1845} & \textbf{88.97} & \textbf{1.57} & 83.14 & \textbf{1.01} \\
\midrule
MCFlow-B & \redx & 62.11 & 0.0624 & 25.32 & 0.1640 & 90.05 & 2.69 & \textbf{84.29} & 1.09 \\
MCFlow-B & \greencheck & \textbf{63.14} & \textbf{0.0611} & \textbf{26.46} & \textbf{0.1572} & \textbf{90.17} & \textbf{3.12} & 83.56 & \textbf{1.20} \\
\midrule
MCFlow-L & \redx & 63.86 &  \textbf{0.0546} & 26.63 & 0.1499 &  \textbf{90.17} & 3.21 & \textbf{84.67} & \textbf{1.33} \\
MCFlow-L & \greencheck &  \textbf{64.08} & 0.0561 &  \textbf{27.16} &  \textbf{0.1401} & 90.07 &  \textbf{3.32} & 83.67 & 1.24 \\
\bottomrule
\end{tabular}
}
\end{table}
\textbf{Ablation study.}
We perform an ablation study on single-sample CSP to evaluate the effect of noisy guidance.
\cref{tab:ng_ablation} shows that applying noisy guidance generally improves the match rate and RMSE on CSP across MP-20 and MPTS-52, indicating enhanced structural fidelity.

\begin{wrapfigure}{r}{0.41\textwidth}
    \centering
    \vspace{-10pt}
    \includegraphics[width=0.41\textwidth]{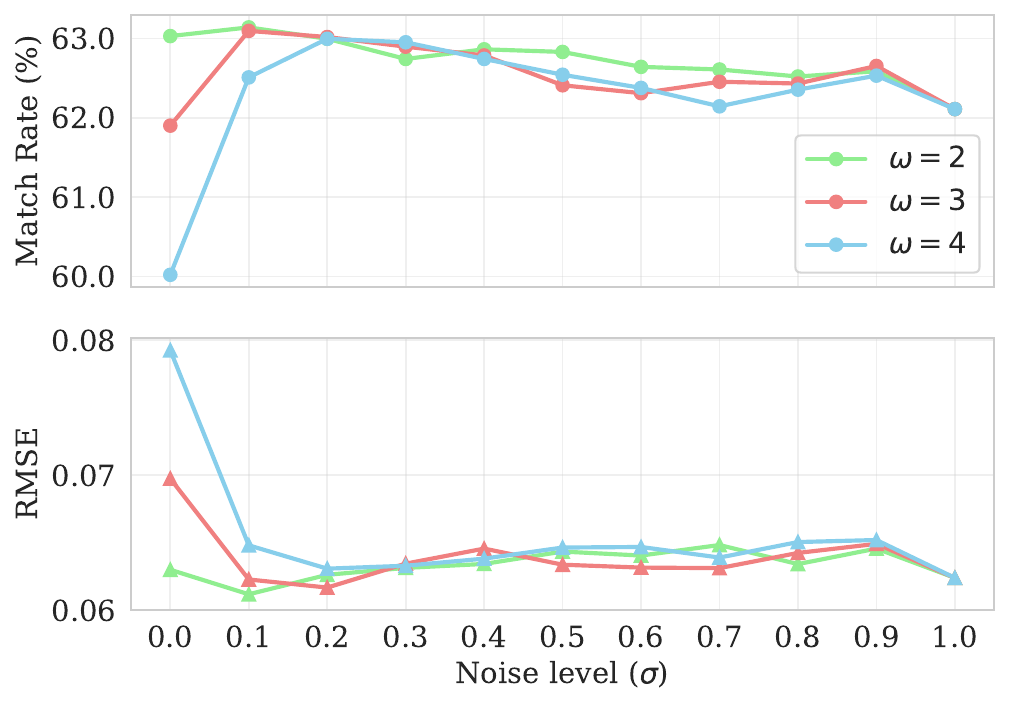}
    \vspace{-10pt}
    \caption{
    \textbf{Sensitivity analysis of noisy guidance.}
    Single-sample CSP performance on MP-20 depending on the atom type noise level $\sigma$ and guidance scale $\omega$.
    }
    \label{fig:ng_sweep}
\end{wrapfigure}

\paragraph{Sensitivity to noise level and guidance scale.}
\cref{fig:ng_sweep} analyzes the sensitivity of noisy guidance to the atom-type noise level $\sigma$ and guidance scale $\omega$ on single-sample CSP.
We observe that introducing a small amount of noise ($\sigma>0$) consistently improves both match rate and RMSE compared to the noiseless setting.
Performance peaks at low noise levels $\sigma= 0.1$, while larger noise gradually degrades structural accuracy.
While the method remains robust across a range of guidance scales, we observe that $\omega=2.0$ consistently yields strong performance.

\clearpage
\section{Model, Training, and Inference Configurations}
\label{app:config}

We provide model architecture (\cref{tab:model_config}), training (\cref{tab:training_configs}), and inference (\cref{tab:inference_configs}) configurations.

\begin{table}[h]
\centering
\caption{MCFlow model configurations.}
\label{tab:model_config}
\begin{tabular}{lccc}
\toprule
Model scale & Small (S) & Base (B) & Large (L) \\
\midrule
\# Parameters & 34M & 134M & 464M \\
Hidden dimension & 384 & 768 & 1024 \\
Transformer layers & 12 & 12 & 24 \\
Attention heads & 6 & 12 & 16 \\
\bottomrule
\end{tabular}
\end{table}

\textbf{Training settings.}
We train all models for a maximum of 10{,}000 epochs using the AdamW optimizer~\citep{loshchilov2017decoupled}
with a constant learning rate of $1\times10^{-4}$ and no weight decay.
To balance learning dynamics across different modalities, we set the loss weights
to 0.5 for atom types, 2.0 for fractional coordinates, and 1.0 for lattice parameters.
The batch size is set to 256.
For computational efficiency, all models are trained using Bfloat16 (BF16) mixed precision.
To ensure training stability, we apply gradient clipping with a maximum norm of 10.0
and clip flow times $t,s$ to $0.9$ to mitigate numerical instability near the boundaries~\citep{campbell2024generative}.
Additionally, we maintain an exponential moving average (EMA) of the model weights
with a decay rate of 0.9999.
The final model is selected by taking the latest epoch among the top three checkpoints ranked by total validity (structural and compositional) on the validation set.
The LogNormal prior parameters for lattice lengths are fitted to the training data after dividing each length by $N^{1/3}$ (\cref{tab:training_configs}).

\begin{table}[h]
\centering
\caption{MCFlow training configurations.}
\label{tab:training_configs}
\begin{tabular}{ll}
\toprule
Hyperparameter & Value \\
\midrule
Optimizer & AdamW \\
Learning rate & $1 \times 10^{-4}$ \\
Weight decay & 0.0 \\
Batch size & 256 \\
Max epochs & 10,000 \\
Precision & BF16 mixed precision \\
Gradient clipping & max norm 10.0 \\
EMA decay & 0.9999 \\
Flow time clipping & $t, s \le 0.9$ \\
Loss weights $(A/F/L^{l}/L^{a})$ & 0.5 / 2.0 / 1.0 / 1.0 \\
LogNormal prior $(\mu, \sigma)$ MP-20 & $(0.847, 0.973, 1.251)$, $(0.243, 0.216, 0.289)$ \\
LogNormal prior $(\mu, \sigma)$ MPTS-52 & $(0.788, 0.972, 1.255)$, $(0.249, 0.206, 0.292)$ \\
\bottomrule
\end{tabular}
\end{table}

\textbf{Inference settings.}
During sampling, we solve the probability flow ODE for continuous variables
and simulate the CTMC for discrete variables using 500 integration steps to
generate crystals.
To improve sample fidelity in single-sample CSP and structure-conditioned
atom type generation, we apply noisy guidance with a guidance scale
$\omega=2.0$ and a noise level $\sigma=0.1$.
In contrast, for 20-sample generation, we disable noisy guidance to better
capture the multimodal nature of the distribution.
The number of atoms is sampled from the empirical distribution of atom counts
in the training set.

\begin{table}[h]
\centering
\caption{MCFlow inference configurations.}
\label{tab:inference_configs}
\begin{tabular}{ll}
\toprule
Hyperparameter & Value \\
\midrule
Integration steps & 500 \\
Guidance scale $\omega$ & 2.0 (single-sample)\\
Noise level $\sigma$ & 0.1 (single-sample)\\
\bottomrule
\end{tabular}
\end{table}

\textbf{Computational resources.}
All experiments for the Small and Base models were conducted on a single NVIDIA GeForce RTX 3090 GPU,
while experiments for the Large model were conducted on a single NVIDIA A40 GPU.
Training the Base model takes approximately 2 days on MP-20 and 4 days on MPTS-52.

\newpage
\section{Evaluation Metrics}
\label{app:metrics}

\textbf{Crystal structure prediction.}
We compare generated structures against ground-truth test structures using \texttt{StructureMatcher} from \texttt{pymatgen}~\citep{ong2013python}.
The Match Rate (MR) is the fraction of generated candidates that match their corresponding ground-truth structures under symmetry-aware matching with tolerances \texttt{stol}=0.5, \texttt{ltol}=0.3, and \texttt{angle\_tol}=10$^\circ$.
For matched pairs only, we compute the Root Mean Square Error (RMSE) of atomic positions, normalized by $(V/N)^{1/3}$, where $V$ is the unit cell volume and $N$ is the number of atoms.

\textbf{\emph{De novo} generation.}
We evaluate \emph{de novo} generation using LeMat-GenBench~\citep{lemat2025genbench}, a standardized benchmark that provides a unified evaluation pipeline.
We sample 2{,}500 structures and apply the LeMat-GenBench validity filter, which checks CIF readability, interatomic distances ($d_{ij} \geq 0.5\,(0.7 + r_i + r_j)$~\AA{} using pymatgen atomic radii), density and lattice-parameter bounds, space-group determinability, and charge neutrality (oxidation-state analysis, tolerance $0.1$). Validity is reported as the pass rate.

Thermodynamic stability is assessed via $E_{\text{hull}}$ using an ensemble of three MLIPs (MACE-MP~\citep{batatia2024macemp}, UMA~\citep{wood2025uma}, Orb-v3~\citep{rhodes2025orbv3}). Each MLIP relaxes the structure and constructs its own self-consistent convex hull from LeMat-Bulk~\citep{siron2025lematbulk}, avoiding mixed MLIP--DFT inconsistencies. We report the ensemble mean and standard deviation of $E_{\text{hull}}$, formation energy ($E_f$), and relaxation RMSD; structures with $E_{\text{hull}} \le 0$ are \emph{stable} and those with $0 < E_{\text{hull}} \le 0.1$~eV/atom are \emph{metastable}.

Novelty and uniqueness use direct structure comparison via pymatgen's \texttt{StructureMatcher} with a stricter \texttt{ltol}=0.1 than the CSP matching above (\texttt{ltol}=0.3), following the LeMat-GenBench convention. S.U.N.\ and MS.U.N.\ apply a stable~$\to$~unique~$\to$~novel funnel to the submitted set, with MS.U.N.\ restricted to the strict metastable range above. All count-based percentages (Valid, Unique, Novel, Stable, Metastable, S.U.N., MS.U.N.) are reported as count over the $2{,}500$ submitted structures, following the LeMat-GenBench leaderboard convention; this caps each percentage above by the validity rate.

\textbf{Atom type generation.}
We report compositional validity (Comp) following \citet{xie2021crystal}: a generated composition is valid if its atomic species satisfy SMACT-based charge-neutrality and electronegativity rules. Match Rate is computed as in CSP, following \citet{jiao2023crystal}.

\textbf{Ablation study.}
Ablation experiments (\cref{tab:ablation,tab:ordering_ablation,tab:unified_ablation,tab:coupled_ablation}) report MR and RMSE for CSP following \citet{jiao2023crystal}, and structural and compositional validity for DNG following \citet{xie2021crystal}: structural validity is the fraction of structures with all interatomic distances above $0.5$~\AA{}, and compositional validity is the fraction of compositions satisfying SMACT-based charge-neutrality and electronegativity rules.

\textbf{Polymorph-aware evaluation.}
We additionally report METRe and cRMSE on a polymorph-aware MP-20 split, both introduced by \citet{martirossyan2025metre}. Standard match rate uses one-to-one pairing between generated and reference structures, which fails when a composition admits multiple stable polymorphs. METRe (Match Everyone to REference) instead compares every generated structure against every reference and counts a reference as matched whenever at least one generated structure falls within tolerance; the METRe rate is the fraction of reference structures that find at least one match. The corrected RMSE (cRMSE) penalizes unmatched references by assigning them an error equal to the matching tolerance \texttt{stol} (with $\texttt{stol}=0.5$, the value used in our CSP evaluation):
\begin{equation}
\mathrm{cRMSE} = \frac{\sum_{i \in \mathrm{matched}} \mathrm{RMSE}_i + \texttt{stol} \cdot (N_{\mathrm{test}} - N_{\mathrm{matched}})}{N_{\mathrm{test}}},
\end{equation}
combining coverage (METRe) and precision (RMSE) into a single score.

\newpage
\section{Additional Results}
\label{app:exp}

\begin{figure}[t]
    \centering
    \includegraphics[width=\textwidth]{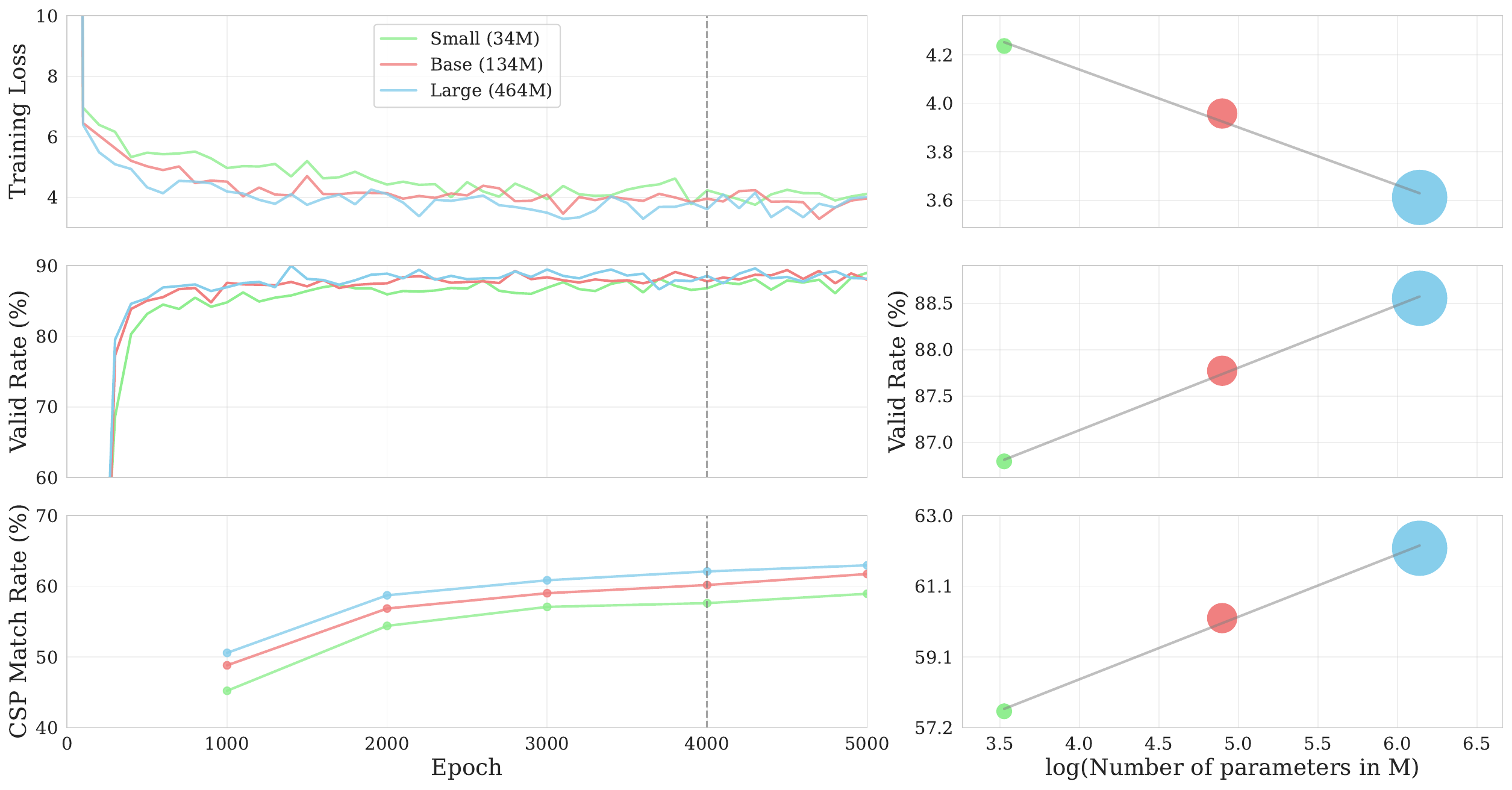}
    \vspace{-15pt}
    \caption{
    \textbf{Scaling behavior of MCFlow across model sizes.}
    Larger MCFlow variants consistently improve training loss, DNG validity, and CSP match rate.
    Training loss, DNG valid rate and CSP match rate over training epochs for small, base, and large models (left).
    Correlation between model size and training loss, validity, and CSP match rate at epoch 4000 (right).
    }
    \label{fig:scaling}
\end{figure}

We provide computational cost analysis (\cref{tab:compute}), polymorph-aware METRe evaluation (\cref{tab:metre}), template-conditioned CSP results (\cref{tab:csp_template}), atom ordering and permutation ablations (\cref{tab:ordering_ablation}), scaling behavior (\cref{fig:scaling}), filtered vs.\ unfiltered match rate (\cref{tab:unfiltered_mr}), DNG on MPTS-52 (\cref{tab:dng_mpts52}), unified vs.\ task-specific training (\cref{tab:unified_ablation}), and decoupled time axes ablation and comparison with adapted DNG baselines (\cref{tab:coupled_ablation}).

\begin{table}[h]
\centering
\caption{
\textbf{Computational cost comparison.}
MCFlow-B achieves faster per-step training and inference than the baselines despite more parameters, because Transformer dense operations are more hardware-efficient than E(3)-equivariant GNN message passing.
Combined with faster metric saturation (\cref{fig:integration_steps}), MCFlow achieves lower total inference cost.
We report wall-clock training time per epoch and inference time per integration step, measured on a single NVIDIA RTX 3090.
}
\label{tab:compute}
\begin{tabular}{lcccc}
\toprule
Model & Params & Batch & Train (1 Epoch) & Infer (1 Step) \\
\midrule
DiffCSP  & 12.3M  & 256 & 26.0s   & 111.1ms \\
FlowMM   & 28.3M  & 256 & 52.11s  & 240.7ms \\
MCFlow-S & 33.7M  & 256 & 8.54s   & 23.9ms  \\
MCFlow-B & 134.0M & 256 & 22.27s  & 68.7ms  \\
MCFlow-L & 464.6M & 128 & 77.68s  & 113.5ms \\
\bottomrule
\end{tabular}
\end{table}

\begin{table}[h]
\centering
\caption{
\textbf{Polymorph-aware CSP evaluation using METRe~\citep{martirossyan2025metre}.}
MCFlow achieves the highest METRe under polymorph and original splits, indicating robustness to evaluation methodology.
METRe (Match Everyone to REference) computes the fraction of reference structures that find at least one matching generated structure; this corrects the standard match rate, whose one-to-one pairing underestimates models that correctly generate multiple stable polymorphs per composition.
}
\label{tab:metre}
\begin{tabular}{llccc}
\toprule
Split & Model & METRe (\%) $\uparrow$ & RMSE $\downarrow$ & cRMSE $\downarrow$ \\
\midrule
\multirow{4}{*}{Polymorph}
& DiffCSP  & 53.1 & 0.084 & 0.279 \\
& FlowMM   & 65.2 & 0.079 & 0.226 \\
& OMatG    & 70.5 & \textbf{0.056} & \textbf{0.187} \\
\cmidrule(lr){2-5}
& \textbf{MCFlow-B} & \textbf{70.7} & 0.068 & 0.195 \\
\midrule
\multirow{6}{*}{Original}
& DiffCSP  & 58.8 & 0.064 & 0.244 \\
& FlowMM   & 67.0 & 0.067 & 0.210 \\
& OMatG    & 66.0 & \textbf{0.058} & 0.208 \\
\cmidrule(lr){2-5}
& MCFlow-S & 62.3 & 0.137 & 0.274 \\
& MCFlow-B & 69.7 & 0.070 & 0.200 \\
& \textbf{MCFlow-L} & \textbf{70.9} & 0.060 & \textbf{0.190} \\
\bottomrule
\end{tabular}
\end{table}

\begin{table}[t]
\centering
\caption{
\textbf{Template-conditioned crystal structure prediction on MP-20 and MPTS-52.}
MCFlow outperforms the template-conditioned baselines DiffCSP++ and SGFM on both datasets.
Evaluation under structural templates extracted from the target structure; for MCFlow, MR and RMSE are computed on generated samples matching the target template.
}

\label{tab:csp_template}
\begin{tabular}{lccccc}
\toprule
\multirow{2}{*}{Model} & \multicolumn{2}{c}{MP-20} & \multicolumn{2}{c}{MPTS-52} \\
\cmidrule(lr){2-3} \cmidrule(lr){4-5}
& MR (\%) $\uparrow$ & RMSE $\downarrow$ & MR (\%) $\uparrow$ & RMSE $\downarrow$ \\
\midrule
DiffCSP++    & 80.27 & 0.0295 & 46.29 & 0.0896 \\
SGFM         & 82.74 & 0.0288 & 51.79 & 0.0827 \\
\midrule
MCFlow-S  & 79.21 & 0.0238 & 73.91 & 0.0415 \\
MCFlow-B  & 85.31 & 0.0210 & \textbf{77.61} & \textbf{0.0361} \\
MCFlow-L  & \textbf{85.96} & \textbf{0.0201} & 74.63 & 0.0386 \\
\bottomrule
\end{tabular}
\end{table}
\begin{table}[t]
\centering
\caption{
\textbf{Impact of atom ordering and permutation strategies on CSP and DNG.}
Electronegativity ordering outperforms atomic number ordering due to the direct correspondence with cation-anion roles exploited by the DiT.
Full $N!$ random permutation substantially degrades performance, as the permutation space overwhelms learning capacity.
All variants use MCFlow-B on MP-20.
}
\label{tab:ordering_ablation}
\begin{tabular}{lcccc}
\toprule
\multirow{2}{*}{Ordering} & \multicolumn{2}{c}{CSP} & \multicolumn{2}{c}{DNG} \\
\cmidrule(lr){2-3} \cmidrule(lr){4-5}
& MR (\%) $\uparrow$ & RMSE $\downarrow$ & Struct (\%) $\uparrow$ & Comp (\%) $\uparrow$ \\
\midrule
Atomic Number    & 59.81 & 0.0951 & 99.30 & 88.09 \\
Rand. Perm.      & 54.25 & 0.1184 & \textbf{99.45} & 86.12 \\
\textbf{Electronegativity} & \textbf{63.14} & \textbf{0.0611} & 99.44 & \textbf{89.85} \\
\bottomrule
\end{tabular}
\end{table}

\newpage
\begin{table}[t]
\centering
\caption{
\textbf{Filtered and unfiltered CSP match rates on MP-20 and MPTS-52.}
MCFlow-L consistently outperforms all baselines under both evaluation schemes.
The unfiltered MR is included to capture physically real structures ($\sim$10\% in MP-20) that fail SMACT validity rules.
Performance is reported as filtered / unfiltered MR using single-sample generation with noisy guidance. OMatG corresponds to the LinearODE variant.
}
\label{tab:unfiltered_mr}
\begin{tabular}{lcccc}
\toprule
\multirow{2}{*}{Model} & \multicolumn{2}{c}{MP-20} & \multicolumn{2}{c}{MPTS-52} \\
\cmidrule(lr){2-3} \cmidrule(lr){4-5}
& MR (\%) $\uparrow$ & RMSE $\downarrow$ & MR (\%) $\uparrow$ & RMSE $\downarrow$ \\
\midrule
DiffCSP  & 52.51 / 57.82 & 0.0600 / 0.0627 & 14.29 / 15.79 & 0.1489 / 0.1533 \\
FlowMM   & 59.98 / 66.22 & 0.0629 / 0.0661 & 20.28 / 22.29 & 0.1486 / 0.1541 \\
OMatG    & 63.75 / 69.83 & 0.0720 / 0.0741 & 25.15 / 27.38 & 0.1931 / 0.1970 \\
\midrule
MCFlow-S & 56.10 / 61.78 & 0.1269 / 0.1321 & 22.88 / 24.74 & 0.1846 / 0.1920 \\
MCFlow-B & 63.14 / 69.23 & 0.0611 / 0.0663 & 26.46 / 28.77 & 0.1577 / 0.1610 \\
\textbf{MCFlow-L} & \textbf{64.08 / 70.38} & \textbf{0.0561 / 0.0592} & \textbf{27.16 / 29.58} & \textbf{0.1401 / 0.1497} \\
\bottomrule
\end{tabular}
\end{table}

\begin{table}[t]
\centering
\caption{
\textbf{De novo generation results on the MPTS-52 benchmark.}
MCFlow achieves the highest compositional validity and competitive structural validity on MPTS-52.
The MPTS-52 dataset contains 100.00\% structurally valid and 88.24\% compositionally valid structures.
\textbf{Bold} and \underline{underline} indicate the best and second-best results, respectively.
}

\label{tab:dng_mpts52}
\begin{tabular}{lcc}
\toprule
Model & Structural Validity (\%) $\uparrow$ & Compositional Validity (\%) $\uparrow$ \\
\midrule
DiffCSP      & 67.47 & 55.80 \\
\midrule
SymmCD       & 87.11 & 78.18 \\
DiffCSP++    & \textbf{99.87} & 77.52 \\
SGEquiDiff   & 97.79 & 79.83 \\
\midrule
MCFlow-S   & 96.73 & 84.62 \\   
MCFlow-B   & \underline{98.34} & \textbf{85.57} \\   
MCFlow-L   & 97.34 & \underline{85.21} \\   
\bottomrule
\end{tabular}
\end{table}
\begin{table}[t]
\centering
\caption{
\textbf{CSP-only vs.\ unified training on MP-20.}
CSP-only ($t=1$ during training) and unified training yield nearly identical MR, indicating no destructive interference.
The unified model additionally supports DNG, ATG, and masked ATG without retraining, and noisy guidance (NG) provides a further gain only available through decoupled time training.
}
\label{tab:unified_ablation}
\begin{tabular}{lcc}
\toprule
Method & MR (\%) $\uparrow$ & RMSE $\downarrow$ \\
\midrule
CSP-only   & 61.92 & 0.0638 \\
MCFlow-B w/o NG          & 62.11 & 0.0624 \\
\textbf{MCFlow-B}        & \textbf{63.14} & \textbf{0.0611} \\
\bottomrule
\end{tabular}
\end{table}

\begin{table}[t]
\centering
\caption{
\textbf{Decoupled time axes ablation and comparison with adapted DNG baselines on MP-20 CSP.}
Decoupled time axes are essential: coupling $t{=}s$ substantially degrades CSP performance. DNG-trained DiffCSP and FlowMM, adapted to CSP by overwriting composition with ground-truth values at every integration step, also achieve substantially lower MR due to their coupled noise schedules.
}
\label{tab:coupled_ablation}
\begin{tabular}{lcc}
\toprule
Method & MR (\%) $\uparrow$ & RMSE $\downarrow$ \\
\midrule
DiffCSP DNG       & \pz3.76 & 0.3396 \\
FlowMM DNG        & 27.01 & 0.1947 \\
\midrule
MCFlow-B coupled ($t{=}s$) & 45.58 & 0.1499 \\
\textbf{MCFlow-B} & \textbf{63.14} & \textbf{0.0611} \\
\bottomrule
\end{tabular}
\end{table}

\clearpage
\newpage
\section{Visualizations of Samples}
\label{app:visualizations}

\begin{figure}[h]
    \centering
    \includegraphics[width=\textwidth]{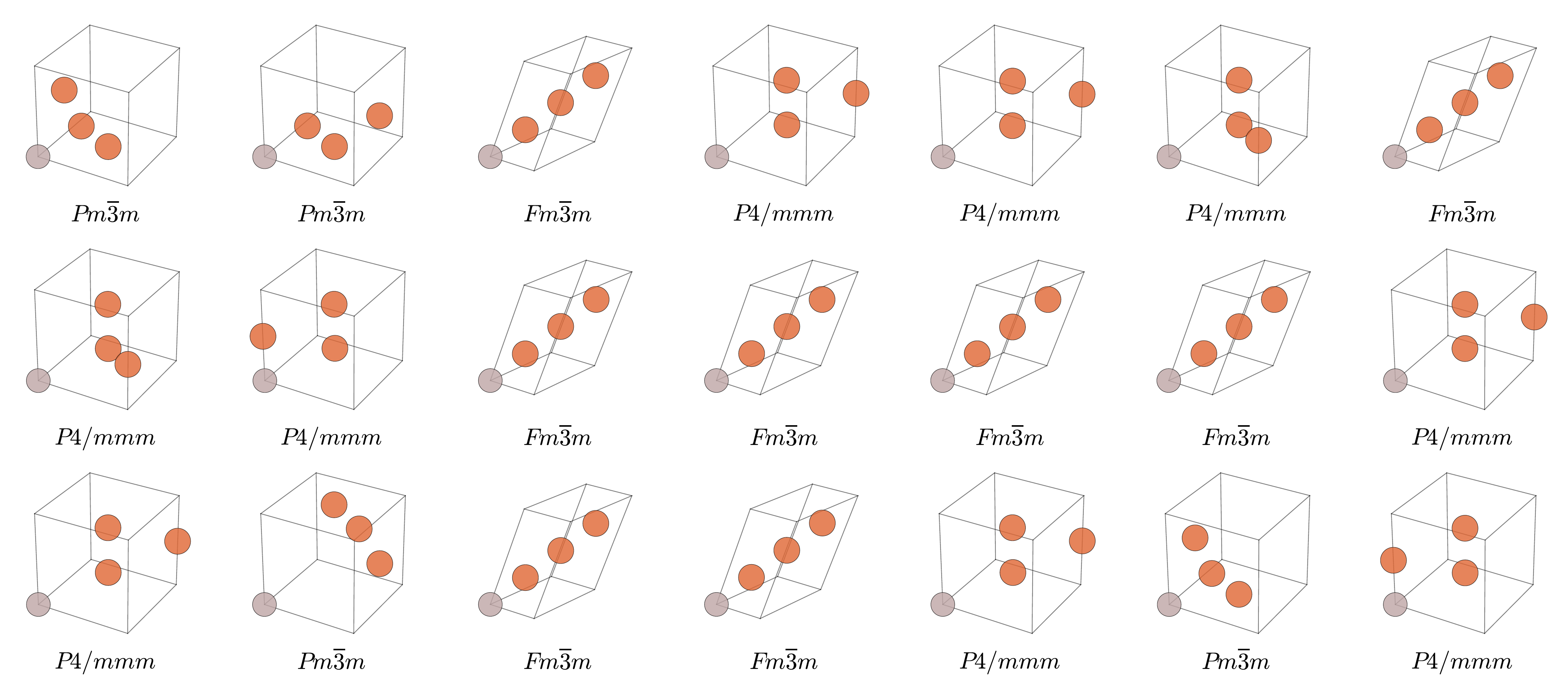}
    \vspace{-15pt}
\caption{
\textbf{Crystal structure prediction.}
MCFlow explores multiple crystal structures with different space groups conditioned on a single composition, illustrated for the alloy AlFe$_3$ (top-left, target crystal).
}

    \label{fig:csp}
\end{figure}

\begin{figure}[h]
    \centering
    \includegraphics[width=\textwidth]{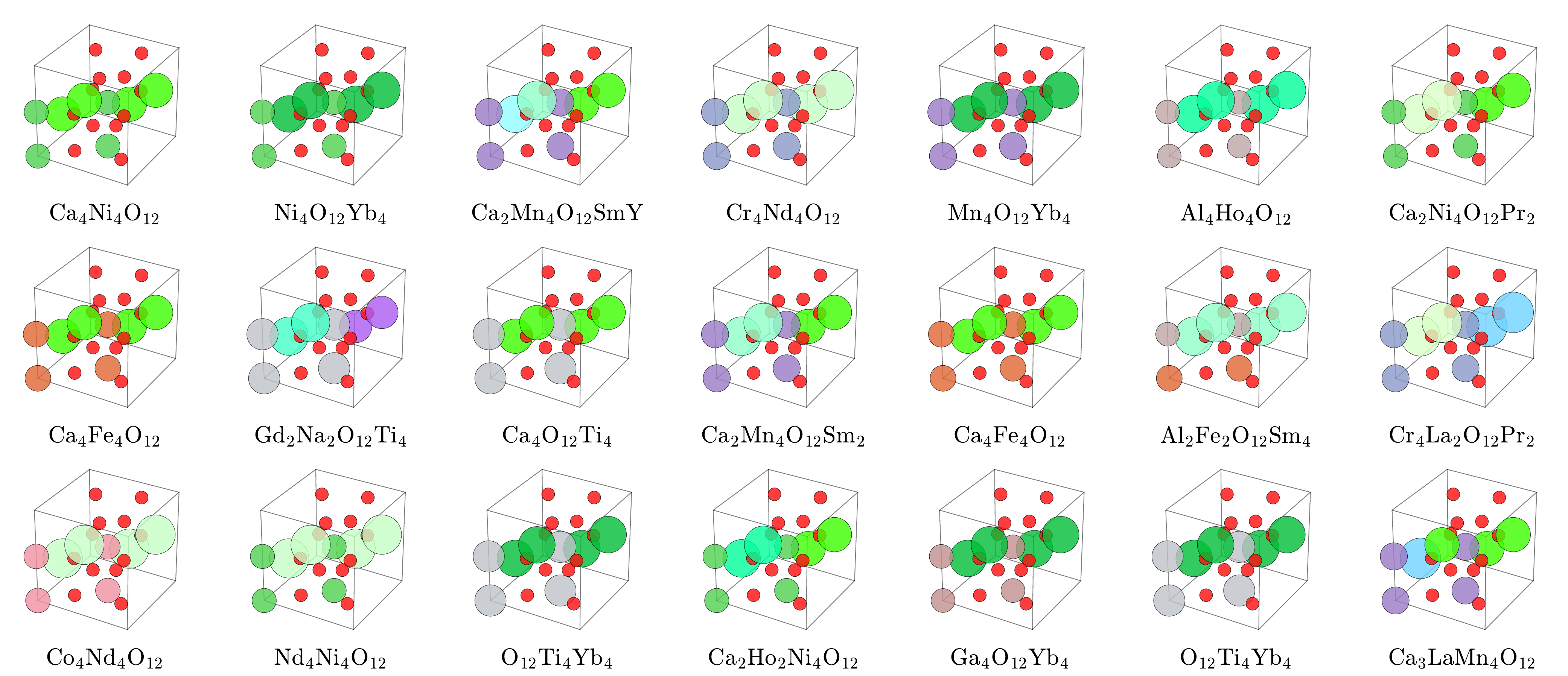}
    \vspace{-15pt}
    \caption{
    \textbf{Atom type generation.}
    MCFlow captures multimodal compositional distributions conditioned on a structure, generating diverse compositions for a given perovskite structure of Ca$_4$Ni$_4$O$_{12}$ (top-left, target crystal).
    }
    \label{fig:atg}
\end{figure}

\begin{figure}[h]
    \centering
    \includegraphics[width=\textwidth]{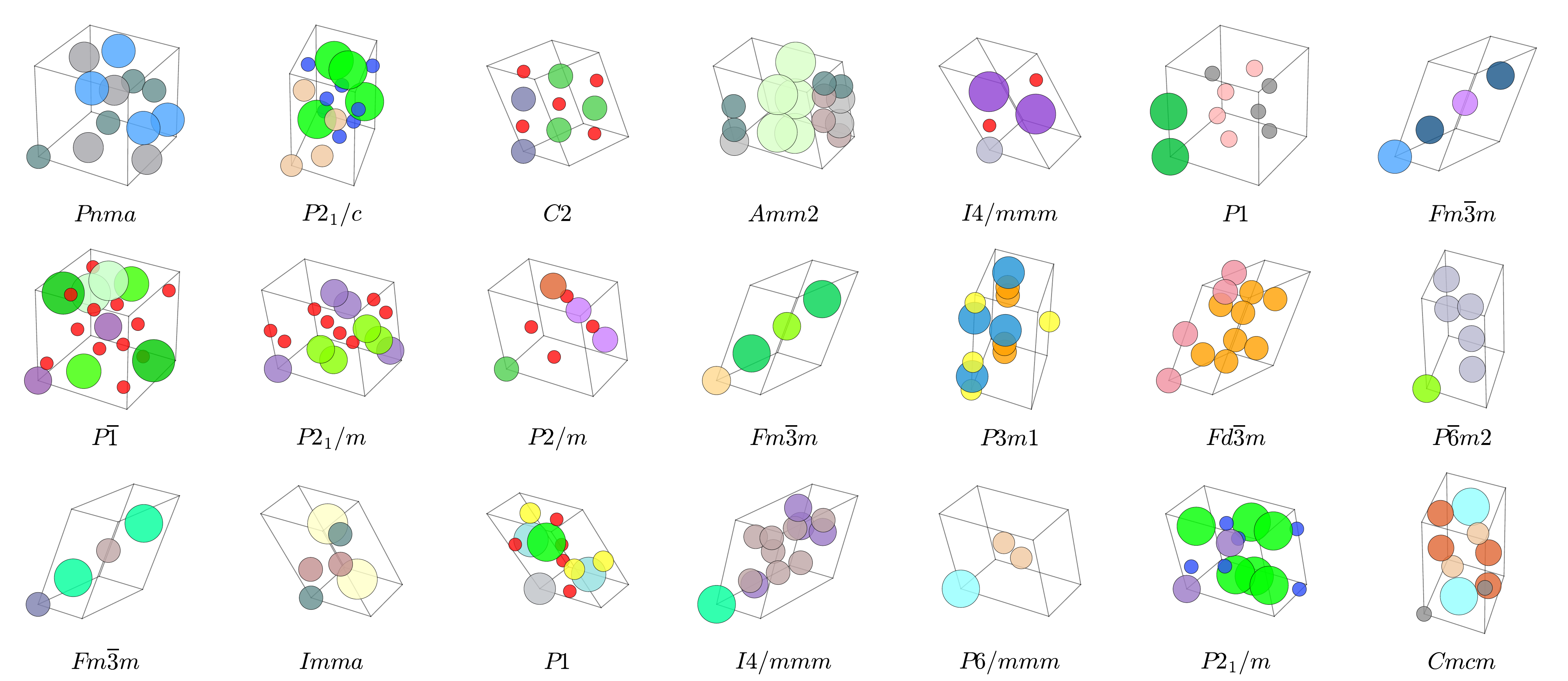}
    \vspace{-15pt}
    \caption{
    \textbf{De novo generation.}
    MCFlow jointly generates diverse atom types and crystal structures.
    }
    \label{fig:dng}
\end{figure}

We visualize qualitative samples from MCFlow across the three tasks: crystal structure prediction (\cref{fig:csp}), atom type generation (\cref{fig:atg}), and \emph{de novo} generation (\cref{fig:dng}).


\end{document}